\pdfoutput=1
\documentclass[11pt]{article}

\usepackage{acl}

\usepackage{times}
\usepackage{latexsym}
\usepackage{graphicx}
\usepackage{tabularx}
\usepackage{booktabs}
\usepackage{afterpage}
\usepackage[T1]{fontenc}
\usepackage{tablefootnote}
\usepackage{tabularx}

\usepackage[utf8]{inputenc}

\usepackage{microtype}
\usepackage{numprint}

\usepackage{inconsolata}
\usepackage[nameinlink]{cleveref}
\usepackage{subcaption}
\usepackage{svg}

\title{Word Boundary Information Isn't Useful for Encoder Language Models}

\author{ 
Edward Gow-Smith\textsuperscript{\textnormal{1}}, Dylan Phelps\textsuperscript{\textnormal{1}}, Harish Tayyar Madabushi\textsuperscript{\textnormal{2}}, \\
\textbf{Carolina Scarton\textsuperscript{\textnormal{1}}} \and
\textbf{Aline Villavicencio\textsuperscript{\textnormal{1}}}
\\[0.3cm]
\textsuperscript{1}Department of Computer Science, University of Sheffield \\
\textsuperscript{2}Department of Computer Science, University of Bath \\
\texttt{\small egow-smith1@sheffield.ac.uk}
\\
}

\begin{document}
\maketitle
\begin{abstract}
All existing transformer-based approaches to NLP using subword tokenisation algorithms encode whitespace (word boundary information) through the use of special space symbols (such as \#\# or \_) forming part of tokens. These symbols have been shown to a) lead to reduced morphological validity of tokenisations, and b) give substantial vocabulary redundancy. As such, removing these symbols has been shown to have a beneficial effect on the processing of morphologically complex words for transformer encoders in the pretrain-finetune paradigm. In this work, we explore whether word boundary information is at all useful to such models. In particular, we train transformer encoders across four different training scales, and investigate several alternative approaches to including word boundary information, evaluating on two languages (English and Finnish) with a range of tasks across different domains and problem set-ups: sentence classification datasets, NER (for token-level classification), and two classification datasets involving complex words (Superbizarre and FLOTA). Overall, through an extensive experimental setup that includes the pre-training of 35 models, we find no substantial improvements from our alternative approaches, suggesting that modifying tokenisers to remove word boundary information isn't leading to a loss of useful information.

\end{abstract}

\section{Introduction}

Transformer \cite{vaswani2017attention} pretrained language models for NLP, such as BERT \cite{devlin2019bert} and the GPT family \cite{brown2020language,achiam2023gpt}, typically use subword tokenisation algorithms, such as WordPiece \cite{schuster2012japanese}, to process text. Previous work \cite{church2020emerging, park2021morphology} has shown that such methods have limited alignment with word morphology, resulting in worsened downstream performance for various tasks \cite{klein-tsarfaty-2020-getting, bostrom-durrett-2020-byte, pinter-etal-2020-will}. In fact, it has been shown that the morphological validity of tokenisation can be improved by removing all whitespace markers (and hence word boundary (WB) information) from the tokenisers \cite{gow2022improving}. However, the full impact of this modification on downstream performance is unknown, and the question of whether WB information is at all useful to models is as yet unanswered. In this work, we first perform a morphological evaluation of WordPiece and WordPiece$'$, a version which has been modified to have no WB information. We find that WordPiece$'$ significantly improves the alignment with morphological gold standard references. Then, we evaluate WordPiece and WordPiece$'$ as tokenisers on downstream tasks. We also introduce models which modify WordPiece$'$ by including WB information in various ways -- either explicitly through the input or implicitly through the pretraining objective. Much interest recently has been in the scaling laws of language models \cite{kaplan2020scaling, hoffmann2022training}, and a direction towards training larger models. On the other hand, there has been recent work investigating sample-efficient pretraining on datasets of a developmentally plausible size \cite{warstadt2023findings}. In companion to such work, we train our models across four training scales, from approximately 6M params and 250M tokens at the lowest scale to approximately 370M params and 23B tokens at the highest scale. 

\begin{figure}[b!]
\def\svgwidth{\columnwidth}
\centering
\includegraphics[width=\columnwidth]{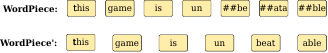}
\caption{\label{fig:wordpiece_prime_tokenisation} Tokenisations generated by WordPiece and WordPiece$'$ for the input sequence ``this game is unbeatable''.}
\end{figure}

Across these scales we pretrain all of our models and evaluate in English on four downstream tasks (comprising 16 datasets): Named Entity Recognition (NER), GLUE, and two tasks involving classifying complex words. We additionally train and evaluate in Finnish across two tasks: NER and Sequence Classification. 

The findings of our work are as follows: (1) we show that modifying WordPiece to remove WB information (giving WordPiece$'$) substantially improves the morphological validity of the resulting tokenisations across English and Finnish; (2) across four training scales, we find that WordPiece$'$ outperforms WordPiece on downstream tasks involving complex words, and gives better performance across most datasets at the lower training scales; (3) we find that none of our methods for including WB information into models, whether implicit or explicit, or through finetuning alone, significantly affects the performance across four downstream tasks and three training scales. Our results indicate that word boundary information isn't providing additional useful information to models, with morphemes being the most important subunit.

\section{Tokenisers}

One particular design choice of subword tokenisers used by transformer models is the addition of prefixes such as ``\_'' and ``\#\#'' in order to encode space information, hence representing word boundaries in languages with spaces between words. Previous work \cite{gow2022improving} has investigated the impact of these prefixes, showing they lead to less morphologically valid tokenisations, and also to a reduced efficiency, since the dual representation of subwords (e.g. ``beat'' and ``\_beat'') gives a vocabulary redundancy (of approximately 9\%). As such, removing these tokens for Unigram \cite{kudo2018subword} and BPE \cite{sennrich2015neural} has been shown to have a beneficial effect on downstream performance for complex word tasks, whilst retaining equivalent performance in general natural language understanding tasks. We refer readers to \citet{gow2022improving} for a full analysis, but here we focus on WordPiece$'$ -- WordPiece modified such that WB information is removed. We train this model and the default on 1 million sentences from Wikipedia for two languages (English and Finnish). We show an example of the tokenisations generated by this compared to the default for English in \Cref{fig:wordpiece_prime_tokenisation}. We perform a morphological evaluation of WordPiece$'$ compared to WordPiece across the two languages, shown in \Cref{table:morphology-comparison}. For English, we use four datasets (LADEC \cite{gagne2019ladec}, MorphoLex \cite{sanchez2018morpholex}, MorphyNet \cite{batsuren2021morphynet}, DagoBERT \cite{hofmann2020dagobert}), and we average across all four (full breakdown in \Cref{table:morphology-comparison-english-full}). For Finnish, we use the subset of MorphyNet. Here, we follow the evaluation standard from \citet{creutz2004morpheme}, reporting precision and F1. Averaging across English and Finnish, we see that WordPiece$'$ gives 14\% shorter sequences, 46\% higher precision, and 34\% higher F1 compared to WordPiece. We also show examples of English tokenisations for WordPiece and WordPiece$'$ in \Cref{table:tokeniser-qualitative-comparison}. In general, we can see that WordPiece generates more meaningful tokenisations, but sometimes they are still of limited morphological validity, as for ``undesirable'' where the prefix is incorrectly split and the base form of the word is lost: we note that WordPiece (like BPE) is a greedy algorithm, meaning it has a tendency to overlengthen the initial token of a word.

\begin{table}[hbt!]
\centering
\resizebox{\linewidth}{!}{%
\begin{tabular}{cc}
WordPiece & WordPiece$'$ \\
\midrule
hyp \#\#ores \#\#po \#\#n \#\#s \#\#iveness & hypo respons iveness \\ 
non \#\#m \#\#ult \#\#ipl \#\#ayer & non multi player \\
over \#\#pr \#\#iced & over price d \\ 
un \#\#icy \#\#cle & uni cycle \\
und \#\#es \#\#ira \#\#ble & und es ira ble \\
\bottomrule
\end{tabular}
}
\caption{\label{table:tokeniser-qualitative-comparison} Some examples of the tokenisations from WordPiece and WordPiece$'$.}
\end{table}

\begin{table}
\centering
\resizebox{\linewidth}{!}{%
\begin{tabular}{ccccccc}
 & \multicolumn{3}{c}{English} & \multicolumn{3}{c}{Finnish} \\ 
 \cmidrule(lr){2-4}\cmidrule(lr){5-7} 
& Len & Precis. & F1  & Len & Precis. & F1 \\
\midrule
WordPiece & 3.29 & 24.8 & 33.8 & 3.21 & 28.3 & 38.9 \\
WordPiece$'$ & 2.75 & \textbf{ 42.6} & \textbf{52.7} & 2.86 & \textbf{34.7} & \textbf{45.0} \\
\bottomrule
\end{tabular}
}
\caption{\label{table:morphology-comparison} Performance of WordPiece and WordPiece$'$ across English and Finnish, showing the average sequence length, precision and F1 score generated following the standard introduced by \citet{creutz2004morpheme}.}
\end{table}

\section{Models}

\begin{figure*}[ht!]
\def\svgwidth{\columnwidth}
\centering
\begin{subfigure}[t]{0.49\textwidth}
\centering
\captionsetup{width=.9\linewidth}
\includegraphics[scale=1]{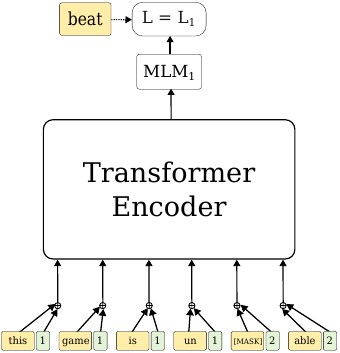}
\caption{\label{fig:explicit_model} Explicit model, where word boundary embeddings are passed in the input.}
\end{subfigure}
\begin{subfigure}[t]{0.49\textwidth}
\centering
\captionsetup{width=.9\linewidth}
\includegraphics[scale=1]{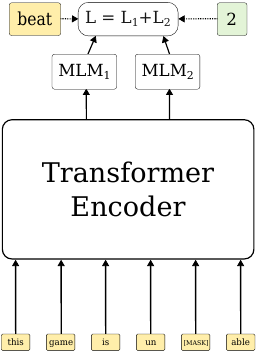}
\caption{\label{fig:implicit_model} Implicit model, with an additional MLM head for predicting word boundaries.}
\end{subfigure}
\caption{\label{fig:network_diagrams} Network diagrams for the modified transformer architectures trained in this work.}
\end{figure*}

The sequences generated by WordPiece$'$ have \textit{no word boundary information}, which means some information is lost when using it to encode sequences. We aim to answer the question of whether such information is at all useful to transformer encoders -- i.e. can it be incorporated in an alternative way to improve performance? We investigate transformer encoders pretrained using the masked language modelling (MLM) task, and then finetuned on downstream tasks (pretrain-finetune paradigm). We then look to include WB information in two ways, either directly as input (both in pretraining and finetuning), or through a modification of the pretraining task.

\subsection{Explicit Model}
\begin{figure}[ht!]
\def\svgwidth{\columnwidth}
\centering
\includegraphics[scale=1.4]{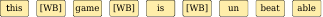}
\caption{\label{fig:wordpiece_prime_spaces} WordPiece$'$ with word boundary tokens.}
\end{figure}

One approach is to include WB information \textit{explicitly} through the input. Naively, we could add WB tokens in the input sequence, shown in \Cref{fig:wordpiece_prime_spaces}. However, this is rather inefficient as it leads to much longer sequences and has been shown to lead to reduced downstream task performance, even when the number of epochs (rather than steps) is matched \cite{gow2022improving}. Nevertheless, we implement this as a baseline.
An alternative, and significantly more efficient, way to include this information is to add ``word boundary embeddings'' to the input, added element-wise with the token embeddings and standard position embeddings, shown in \Cref{fig:explicit_model}. These embeddings are equivalent to the standard position embeddings in being randomly-initialised and then learned through training.

\begin{figure}[ht!]
\def\svgwidth{\columnwidth}
\centering
\includegraphics[scale=1.4]{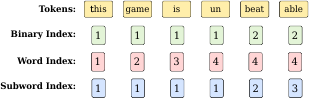}
\caption{\label{fig:word_position_indices} Three alternative indexing methods for the word boundary embeddings.}
\end{figure}

We experiment with three methods for indexing the WB embeddings: \textit{binary index}, \emph{word index}, and \emph{subword index}, shown in \Cref{fig:word_position_indices}. The \textit{word index} is the position of the word the corresponding token belongs to, whereas the \textit{subword index} is the position within the word. These are chosen to align with how the standard position indices work within transformer architectures, but the \textit{binary index} aligns with how standard WordPiece processes word-initial and word-internal tokens, having a value of 1 if a token appears at the start of the word, and a value of 2 otherwise. The \textit{binary index} is also more parameter-efficient, since it only requires an embedding dimension of 2. In fact, for our experiments the \textit{subword index} gives the most new parameters, since even in our English pretraining corpora (Wikipedia and C4) we encounter large chunks of (e.g. Chinese) text with no whitespace, requiring a high embedding dimension.\footnote{We set the embedding dimension at 512, which covers all text encountered for all scales. For the word index, the embedding dimension is set at the max sequence length (256).} 

\subsubsection{Finetuning}
Alongside including WB information at pretraining, we also experiment with pretraining using the default MLM task and architecture, and then passing the WB information during finetuning only, either with binary index WB embeddings, or WB tokens.

\subsection{Implicit Model}
One possible drawback of the explicit approach is the reduced difficulty of the MLM task: passing WB information in the input allows the model to utilise this directly for predicting the masked token, rather than inferring it from context alone. Thus, as an alternative, we modify the architecture with an additional MLM head such that the model has to predict the word boundaries from the input, which we state as \textit{implicitly} using WB information through backpropagation. We show the architecture in \Cref{fig:implicit_model}. In this set-up, we simply sum the losses from the two MLM heads to give the overall loss.\footnote{In preliminary experiments we tried weighting the two losses, but no increase in performance was observed.}

\begin{table}[t!]
\centering  
\resizebox{\columnwidth}{!}{%
\begin{tabular}{lcccccc}
\toprule
 & \multicolumn{2}{c}{\# Articles (M)} & Params (M) & Batch Size & \# GPUs & Steps (k) \\
 & Eng. & Fin. & & & & \\
\midrule
V Low  & 0.1 & 0.1 & 5.8 & 1024 & 1 & 25  \\
Low &  0.5 & 2 & 21.2 & 512 & 1  & 50 \\
High  & 6.5 & 10 & 98.2 & 256 & 1 & 400 \\
V High  & 40 & - & 370.4 & 128 & 4 & 400  \\
\bottomrule
\end{tabular}
}
\caption{\label{table:resource_setups} The four training scales we use to evaluate our models.}
\end{table}

\section{Experiments}

\begin{table*}[ht!]
\centering  
\resizebox{\textwidth}{!}{%
\renewcommand{\arraystretch}{1.2}
\begin{tabular}{lcccccccccccccccc}
  &  \multicolumn{4}{c}{GLUE} &  \multicolumn{4}{c}{NER}  &  \multicolumn{4}{c}{Superbizarre} &  \multicolumn{4}{c}{FLOTA} \\ 
 \cmidrule(lr){2-5}\cmidrule(lr){6-9} \cmidrule(lr){10-13} \cmidrule(lr){14-17}
 & V Low & Low & High & V High & V Low & Low & High & V High & V Low & Low & High & V High & V Low & Low & High & V High \\
\midrule
WordPiece & 54.7 (.6) & 67.7 (1.5) & 77.9 (.4) & 83.1 (.4) & 54.3 (.5) & \textbf{68.9 (.4)} & \textbf{76.9 (.3)} & 81.5 (.4) & 65.7 (.1) & 66.2 (.1) & 67.3 (.1) & 68.6 (.1) & 19.5 (.8) & 31.2 (3.7) & 50.4 (.7) & 55.0 (1.1) \\
WordPiece$'$ & \textbf{56.2 (.4)} & \textbf{69.8 (.5)} & 78.0 (.2) & 83.7 (1.1) & 53.6 (.6) & 68.0 (.5) & 75.7 (.2) & 81.5 (.4) & \textbf{66.9 (.1)} & \textbf{67.6 (.1)} & \textbf{68.4 (.3)} & \textbf{69.5 (.2)} & \textbf{23.6 (.4)} & \textbf{43.1 (.2)} & \textbf{52.3 (.5)} & 55.2 (1.0) \\
\bottomrule
\end{tabular}
}
\caption{\label{table:results-wordpiece-wordpiece-prime} English results across the four tasks and training scales for WordPiece and WordPiece$'$, with standard deviations in parentheses. Results in bold are those better by more than the combined standard deviation ranges.}
\end{table*}

\begin{table*}[ht!]
\centering  
\resizebox{\textwidth}{!}{%
\renewcommand{\arraystretch}{1.2}
\begin{tabular}{lcccccccccccc}
  &  \multicolumn{3}{c}{GLUE} &  \multicolumn{3}{c}{NER}  &  \multicolumn{3}{c}{Superbizarre} &  \multicolumn{3}{c}{FLOTA} \\ 
 \cmidrule(lr){2-4}\cmidrule(lr){5-7} \cmidrule(lr){8-10} \cmidrule(lr){11-13}
 & V Low & Low & High & V Low & Low & High & V Low & Low & High & V Low & Low & High \\
\midrule
WordPiece$'$ & 56.2 (.4) & 69.8 (.5) & 78.0 (.2) & 53.6 (.6) & 68.0 (.5) & 75.7 (.2) & 66.9 (.1) & 67.6 (.1)& 68.4 (.3) &23.6 (.4) & 43.1 (.2) & 52.3 (.5) \\
WordPiece$'$ implicit & 56.2 (.3) & 69.0 (.2) & 77.8 (.8) &  55.3 (.3) & 69.2 (.2) & 75.6 (.4) &  66.9 (.1) & 67.6 (.1) & 68.3 (.1) &  23.5 (1.1) & 45.1 (.8) & 51.8 (1.3) \\
WordPiece$'$ explicit binary & 55.7 (.4) &	70.1 (.2) &	78.4 (.5) &	54.4 (.4) &	68.2 (.8) &	75.3 (.4) & 66.9 (.1) &	67.6 (.1) &	68.2 (.1) &	24.5 (1.7) & 44.5 (.9) & 51.8 (.6) \\
WordPiece$'$ explicit word & 57.2 (.4) & 69.2 (.1) & 78.8 (.3) & 54.9 (.3) & 68.4 (.3) & 75.4 (.4) & 66.8 (.1) & 67.6 (.1) & 68.4 (.1) & 22.3 (.7) & 43.2 (1.0) &51.0 (2.7)\\
WordPiece$'$ explicit subword & 55.6 (.6) &	70.3 (.2) &	78.1 (.4) & 55.0 (.3) &	68.1 (.4) &	75.4 (.3) &	67.0 (.1) &	67.7 (.2) &	68.2 (.2) &	24.3 (1.2) & 38.2 (4.9) & 51.8 (2.8) \\
WordPiece$'$ explicit WB tokens & 55.3 (.6) & 68.7 (.2) & 77.5 (2.0) & 52.4 (.5) & 67.6 (.2) & 74.1 (.2) & 66.6 (.1) & 67.5 (.1) & 68.3 (.2) & 23.3 (1.1) & 43.5 (.2) & 52.3 (.1) \\
WordPiece$'$ explicit f/t WB tokens & 55.1 (.2) & 69.8 (.3) & 76.7 (.5) & 53.6 (.6) & 68.3 (.4) & 75.7 (.2) & - & - & - & 23.4 (1.5) & 43.7 (.7) & 52.5 (.8) \\
WordPiece$'$ explicit f/t binary & 56.2 (.6) & 69.9 (.4) & 77.8 (.4) & 53.6 (.3) & 68.6 (.5) & 75.4 (.4) & 66.9 (.1) & 67.5 (.1) & 68.1 (.4) & 23.4 (1.2) & 43.6 (1.5) & 52.6 (1.3) \\
\bottomrule
\end{tabular}
}
\caption{\label{table:results-wordpiece-prime-space-models} English results across the four tasks and three training scales for WordPiece$'$ and the modified architectures which include word boundary information, with standard deviations in parentheses.}
\end{table*}

\begin{table}[ht!]
\centering  
\resizebox{\columnwidth}{!}{%
\renewcommand{\arraystretch}{1.2}
\begin{tabular}{lcccccc}
  &  \multicolumn{3}{c}{NER} & \multicolumn{3}{c}{SeqClass} \\ 
 \cmidrule(lr){2-4}\cmidrule(lr){5-7}
 & V Low & Low & High & V Low & Low & High \\
\midrule
WordPiece & 72.2 (.2) & 84.2 (.6) & 89.9 (.3) & 73.1 (.2) & 78.7 (.3) & 83.6 (.2)\\
WordPiece$'$ & 73.0 (.6) & 85.0 (.4) & 89.8 (.2) & 73.0 (.6) & 79.0 (.5) & \textbf{84.1 (.3)} \\
\bottomrule
\end{tabular}
}
\caption{\label{table:results-wordpiece-wordpiece-prime-finnish} Finnish results across the three tasks and training scales for WordPiece and WordPiece$'$, with standard deviations in parentheses. Results in bold are those better by more than the combined standard deviation ranges.}
\end{table}

We evaluate the two tokenisers (WordPiece and WordPiece$'$) and our seven explicit and implicit models in the pretrain-finetune paradigm for English and Finnish across three training scales (V Low, Low, High), with an additional scale (V High) for English WordPiece and WordPiece$'$ (unmodified) -- due to the high computational cost of training, we don't train the other models at this scale. Across these scales we vary the number of parameters, batch size, and training steps, shown in \Cref{table:resource_setups}, with further detail in the appendix in \Cref{table:resource_setups_full_english,table:resource_setups_full_finnish}. The first three set-ups for English, and the first two for Finnish, take the training data from Wikipedia, whilst the remaining take data from C4 \cite{raffel2020exploring}. The number of parameters is altered by adjusting the layers, attention heads, and embedding dimension, and a breakdown of this is given in the appendix in \Cref{table:parameters}. We train our models in the manner of RoBERTa \cite{liu2019roberta} (in comparison to BERT, this involves no next sentence prediction, and dynamic masking is performed), and we mask 15\% of tokens. Across all set-ups, we linearly warmup the learning rate to a maximum value of 1e-4, and then linearly decay to 0. We use a sequence length of 256. All training is performed on A100 or H100 GPUs. Training and validation losses for these models are given in the appendix: \Cref{figure:loss_curves_english,figure:loss_curves_finnish}.

For these models, we run an evaluation on four downstream tasks.
The first two tasks focus on natural language understanding across a broad range of domains:

\paragraph{GLUE} We evaluate on 8 GLUE \cite{wang-etal-2018-glue} tasks (excluding the 9th task of WNLI \cite{levesque2012winograd}, following previous work, due to its adversarial nature). These tasks all involve sequence classification, and cover a wide range of domains and set-ups: two single-sentence tasks, three similarity and paraphrase tasks, and three inference tasks. We report the average metric across all tasks. 

\paragraph{NER} We evaluate on three NER datasets from different domains: the English portion of the CoNLL-2003 NER dataset \cite{tjong-kim-sang-de-meulder-2003-introduction}, consisting of sentences taken from the Reuters news corpus \cite{rose-etal-2002-reuters}; the NCBI Disease corpus \cite{dougan2014ncbi}, consisting of PubMed abstracts; and the WNUT2017 Shared Task \cite{derczynski2017results}, with training data taken from Twitter, and test data from YouTube. 

The final two tasks specifically involve morphologically complex words, where we expect more morphologically valid tokenisations to result in improved performance:

\paragraph{Superbizarre} The Superbizarre datasets \cite{hofmann-etal-2021-superbizarre} involve the binary classification of standalone complex words. We take the two topicality datasets: Arxiv, which involves predicting whether a word comes from the Physics or Computer Science subject areas; Reddit, which involves predicting whether a word comes from an entertainment or discussion subreddit. We report the average macro F1 across the two datasets. 

\paragraph{FLOTA} The datasets introduced alongside the FLOTA tokenisation method \cite{hofmann-etal-2022-embarrassingly} involve classifying the title of an Arxiv paper into one of 20 subareas for three subject areas (Computer Science, Maths, Physics). We take the small version of the dataset, with a train set of \numprint{2000} titles per subject area. We report the average macro F1 across the three datasets. 

\subsection{Finnish}
In addition to our experiments on English, we train models on Finnish, to see whether our results are transferable to a morphologically complex language -- one could hypothesise that with greater morphological complexity, word boundary information would be more helpful in disambiguation. We run our experiments on Finnish for WordPiece and WordPiece$'$ across three training scales, and evaluate on two downstream tasks:

\paragraph{NER} We evaluate on the FiNER dataset \cite{ruokolainen2020finnish}, consisting of news articles annotated with six entity classes, reporting macro F1.

\paragraph{Sequence Classification} We look at two sequence classification datasets: the Eduskunta dataset,\footnote{\url{https://github.com/aajanki/eduskunta-vkk}} consisting of ministers' answers to questions from MPs, labelled with the relevant ministry; the FinnSentiment dataset \cite{linden2023finnsentiment}, consisting of sentences from social media labelled with their polarity. We report the accuracy over these two datasets.

\subsection{Finetuning Procedure}
An overview of all datasets is given in \Cref{table:dataset_info}.
We finetune on each dataset by updating all parameters, with the following hyperparameters: batch size 32, max sequence length 128, learning rate of 2e-5, warm-up for 5\% of steps. We evaluate every epoch on the dev set, taking the best-performing epoch. We train five seeds for every model and report the average metric across these. We also remove outliers which lie more than two standard deviations from the mean, or when very low scores suggest the model failed to train.\footnote{This occurs for the following. High: one seed of WordPiece$'$ FLOTA CS (score of 7), one seed of WordPiece$'$ FLOTA Maths (score of 11), one seed of WordPiece$'$ f/t WB tokens (score of 3); V High: one seed of WordPiece$'$ WB tokens CoLA (score of 0), two seeds of WordPiece CoLA (scores of 0 and 8), one seed of WordPiece$'$ STS-B (score of 2), one seed of WordPiece FLOTA CS (score of 4),  one seed of WordPiece FLOTA Maths (score of 3).} For the English NER and Complex Words Datasets, and all Finnish datasets, we train for 20 epochs, but for GLUE we limit it to 10 epochs per dataset due to the relatively high training time.

\begin{figure*}[ht!]
\centering
    \begin{subfigure}[t]{0.24\textwidth}
    \begin{center}
        \resizebox{.99\linewidth}{!}{\includegraphics{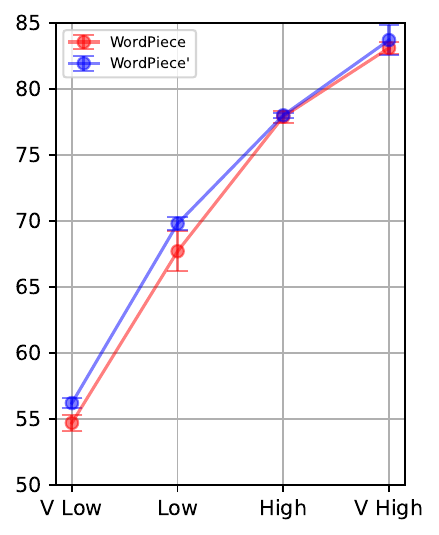}}
    \end{center}
    \caption{GLUE}
    \end{subfigure}
    \begin{subfigure}[t]{0.24\textwidth}
    \begin{center}
       \resizebox{.99\linewidth}{!}{\includegraphics{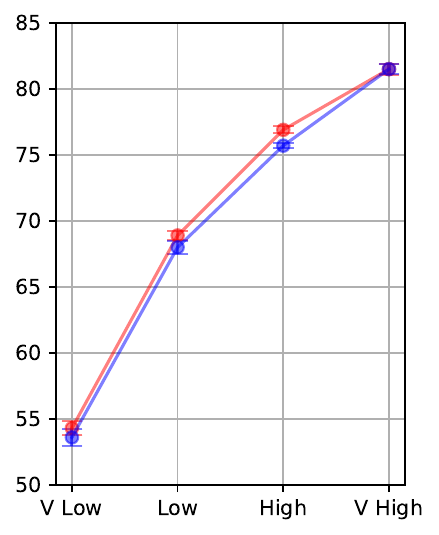}}
    \end{center}
    \caption{NER}
    \end{subfigure}
    \begin{subfigure}[t]{0.24\textwidth}
    \begin{center}
        \resizebox{.99\linewidth}{!}{\includegraphics{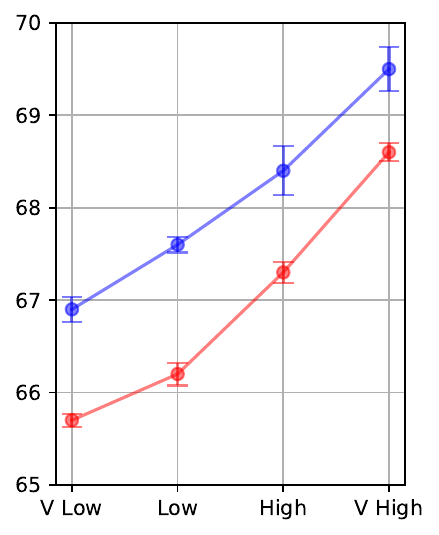}}
    \end{center}
    \caption{Superbizarre}
    \end{subfigure}
    \begin{subfigure}[t]{0.24\textwidth}
    \begin{center}
        \resizebox{.99\linewidth}{!}{\includegraphics{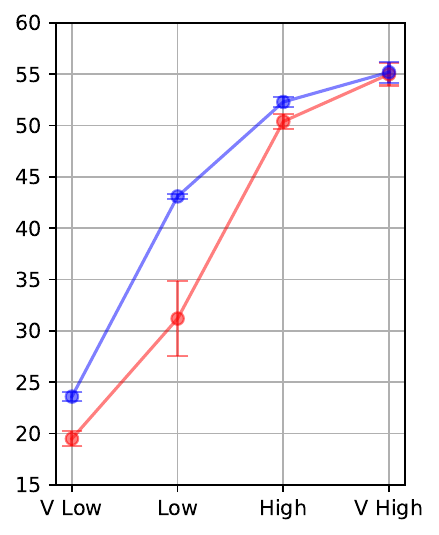}}
    \end{center}
    \caption{\label{figure:results_wordpiece_wordpiece_prime_flota}FLOTA}
    \end{subfigure}
    
    \caption{\label{figure:results_wordpiece_wordpiece_prime} English results for WordPiece and WordPiece$'$ across four training scales and four tasks.}
\end{figure*}

\begin{figure}[ht!]
\centering
    \begin{subfigure}[t]{0.23\textwidth}
    \begin{center}
        \resizebox{.99\linewidth}{!}{\includegraphics{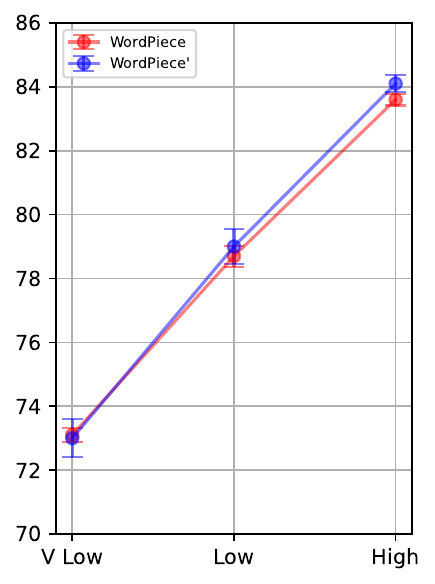}}
    \end{center}
    \caption{SEQ}
    \end{subfigure}
    \begin{subfigure}[t]{0.23\textwidth}
    \begin{center}
       \resizebox{.98\linewidth}{!}{\includegraphics{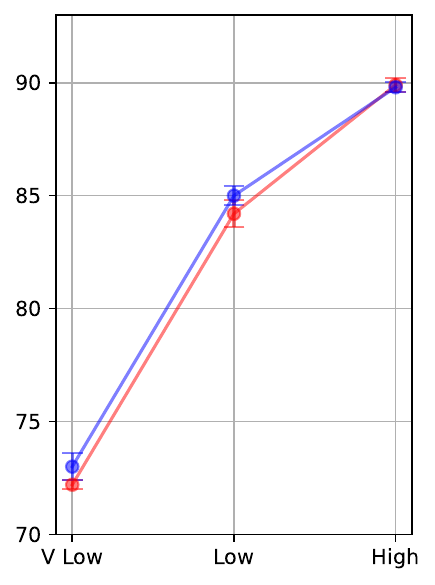}}
    \end{center}
    \caption{NER}
    \end{subfigure}
    \caption{\label{figure:results_wordpiece_wordpiece_prime_finnish} Finnish results for WordPiece and WordPiece$'$ across three training scales and two tasks.}
\end{figure}

\section{Results}
\label{sec:results}

We report our full results across all individual datasets for all models in the appendix (\Cref{table:full_results,table:full_results_finnish}). Here, we look at the overall metrics from the four tasks across the training scales, and present our main findings.
We note that the plots produced  (\Cref{figure:results_wordpiece_wordpiece_prime,figure:results_wordpiece_wordpiece_prime_finnish}, and \Cref{figure:results_wordpiece_extra_loss,figure:results_wordpiece_prime_spaces,figure:results_wordpiece_prime_wb_embeddings,figure:results_wordpiece_prime_ft} in the appendix) are approximately logarithmic in training scale, and we reproduce them using a scale factor on the x-axis in the appendix: \Cref{figure:log_results_wordpiece_wordpiece_prime,figure:log_results_wordpiece_extra_loss,figure:log_results_wordpiece_prime_spaces,figure:log_results_wordpiece_prime_wb_embeddings,figure:log_results_wordpiece_prime_ft}.

Firstly, we compare WordPiece and WordPiece$'$ in \Cref{table:results-wordpiece-wordpiece-prime} and \Cref{figure:results_wordpiece_wordpiece_prime}. On GLUE, we see that WordPiece$'$ performs better than WordPiece across all scales, with a bigger performance difference at the lower scales (+1.5 and +2.1 for the V Low and Low training scales, respectively). We note that at the higher scales, the differences are within two standard deviations of the baseline, so these results are consistent with those by \citet{gow2022improving}. For NER, on the other hand, we find that WordPiece$'$ performs worse than WordPiece across all training scales except V High, where they perform equivalently. Looking at the individual dataset performances (\Cref{table:full_results} in the appendix) we see that the worse performance on WNUT2017 (-2.5 average decrease across scales) accounts for the worse overall NER performance, with the other two datasets giving similar results (apart from at the V Low scale, where WordPiece$'$ performs substantially better on them). This dataset involves tagging ``unusual, previously-unseen entities'', which means morphological composition cannot be leveraged -- we hypothesise that the improved ability of WordPiece$'$ to do this is the cause of the performance drop, due to the futility of composing the meaning of novel surface forms from subunits. Our results on Finnish (\Cref{table:results-wordpiece-wordpiece-prime-finnish} and \Cref{figure:results_wordpiece_wordpiece_prime_finnish}) show no significant performance difference between WordPiece and WordPiece$'$ across the sequence classification and NER tasks, apart from for the High training scale on sequence classification, where WordPiece$'$ outperforms WordPiece.

For the complex word tasks, WordPiece$'$ substantially outperforms WordPiece: averaging across the training scales, we get 1.1 average increase for Superbizarre, and 4.5 average increase for FLOTA. The relative performance difference is most significant for Superbizarre: at the V Low scale, we would require approximately 20 times the training scale for WordPiece to match WordPiece$'$ (\Cref{figure:log_results_wordpiece_wordpiece_prime_superbizarre} in the appendix). In general, we find the performance differences to decrease as the training scale increases, as expected,\footnote{Improved morphological validity should matter less when the model capacity is greater, and when morphologically complex and rare words have been encountered more times during pretraining.} however this effect seems significantly less for Superbizarre, which still has a large performance difference at the V High training scale (+0.9).

Next, we look at the models that attempt to use WB information, with results in \Cref{table:results-wordpiece-prime-space-models}.

Comparing WordPiece$'$ and the \textit{implicit} variant (shown also in \Cref{figure:results_wordpiece_extra_loss} in the appendix), we find that adding the extra loss term gives mixed results across the four tasks and training scales. We do however see that at the V Low and Low training scales, the implicit model improves performance for NER (+1.7 and +1.2, respectively). Since this prediction task is very similar to the finetuning task of token classification, this may explain the effect on performance. The additional MLM head increases the total loss (see \Cref{figure:losses} in the appendix), but when we look at the evaluation accuracies for the two MLM heads (\Cref{fig:wordpiece_prime_eval_losses} in the appendix), we see that the default MLM head has very similar accuracies to the WordPiece$'$ baseline. We also note that for the Very Low training scale, there is a 3.5\% (relative) improvement in default MLM accuracy, which could be contributing to the performance improvement -- in a low resource scenario (both compute and data), the extra prediction task may help to leverage additional information. 

Next, we look at the \textit{explicit} variants. Naively including the WB information through additional tokens leads to decreased performance across all tasks except for FLOTA, where there is no substantial performance difference (\Cref{figure:results_wordpiece_prime_spaces} in the appendix). Overall, these differences are small: around 1 for GLUE, 0.5-2 for NER, 0.1-0.3 for Superbizarre. This is despite a significantly lower MLM loss (approximately 60\%: \Cref{figure:losses} in the appendix) due to the high probability of WB tokens, and the fact that this model trains for around 40\% fewer epochs (\Cref{table:resource_setups_full_english} in the appendix). We next look at the three variants for WB embeddings (see appendix: \Cref{figure:results_wordpiece_prime_wb_embeddings}). Overall, none of these models consistently improve over WordPiece$'$, and the relative performance of the three indexing methods varies with training scale and task. The subword index model has the greatest number of additional parameters, which might explain why this model performs the best overall at the V Low scale. In this setting this model has 2.3\% more parameters than the baseline, compared to 1.1\% for the word index model, and 0.01\% for the binary index model. The model achieves an average performance across the four tasks of 50.5, compared to 50.3 for the other two variants, and 50.1 for the baseline. However, at the Low training scale, this model actually performs worse than the other two variants (61.1 average compared to 62.6 and 62.1 for binary and word, respectively). At the High training scale they all perform equivalently (68.4 average). Since all three indexing methods are encoding equivalent information through trivial transformations, the performance equivalence is perhaps expected.

Finally, we look at two approaches to including WB information during finetuning only (\Cref{figure:results_wordpiece_prime_ft} in the appendix) -- with WB tokens or binary index WB embeddings. We find that neither of these approaches improve over the baseline, with the WB tokens approach performing overall slightly worse: averaged across all training scales and datasets, we get 57.5 for default WordPiece$'$, 57.5  for WordPiece$'$ f/t binary index, and 57.3 for WordPiece$'$ f/t WB tokens. This corroborates the results by \citet{abdou-etal-2022-word}, who find that adding position embeddings after pretraining without them does not lead to improved performance. On average, including the WB embeddings during finetuning decreases training stability (increased standard deviation across seeds).

\section{Discussion}
Overall, we find that \textit{incorporating word boundary information in transformer encoders, either explicitly or implicitly, does not lead to substantial performance improvements}. This suggests that: a) modifying tokenisers such as WordPiece to remove space information does not result in the loss of useful information, b) the default MLM task is sufficient for such models to pretrain effectively. 

The pre-tokenisation step of splitting on whitespace prevents tokens from ever crossing word boundaries, which is perhaps a sufficient restriction.
Our results indicate the importance of a morpheme compared to a word as the key feature which contributes to meaning. 

% In child development studies... 
% The impact of removing word boundaries on ambiguity is undetermined. There will be ambiguous sentences but their prevalence is low. One example is noun compounds such as "greenhouse" vs. "green house". In terms of affixes, there are examples of ambiguity between a suffix of the preceding word and an affix of the following one. An example is...

For English, across all models and training scales, we only see a weak correlation between performance on NER and GLUE -- if we compare the difference compared to WordPiece$'$ for the implicit and explicit models, we find a correlation with Pearson's $\rho = 0.332$.

The Superbizarre task is significantly less affected by model scaling than the other tasks we evaluate on, but much more affected by the choice of tokeniser. This suggests that morphologically valid tokenisation is vital for generating good representations of complex words in the absence of context. This task is also less likely to be dependent on spurious correlations (annotation artefacts) in the data.

All of our models at the High and V High training scales outperform the dev results reported by \citet{hofmann-etal-2022-embarrassingly} on the FLOTA ArXiv-S datasets using their tokenisation method. We hypothesise this is likely an effect of hyperparameters, e.g. we use a batch size of 32 rather than their 64, and we use a learning rate scheduler with warm-up, whereas they do not.

\section{Related Work}
This work aligns with other works that aim to improve the morphological validity of subword tokenisers: \citet{westhelle2022impact} introduce Morphologically Informed Segmentation (MIS), a tokeniser based on Morfessor for Portuguese; \citet{hofmann-etal-2022-embarrassingly} introduce Few Longest
Token Approximation (FLOTA), which preserves the morphology of complex words without necessarily keeping all the characters. \citet{jimenez-gutierrez-etal-2023-biomedical} introduce a tokeniser for the biomedical domain that is better aligned with morpheme segmentation, and then train their BioVocabBERT model using it.
There has also been work looking at the impact of how subword tokens are marked, either with word-initial or word-final prefixes \cite{jacobs2022lost}.

There is previous work which has passed additional position indices to transformer models. \citet{jia2021png} introduce a model for neural text-to-speech called PnG BERT which uses word-position embeddings to provide alignment between phonemes and graphemes at the word level. In NLP, \citet{bai2020segatron} introduce Segatron, a model which modifies the Transformer-XL \cite{dai2019transformer} with two additional position embeddings: a sentence index and a paragraph index. They also apply the same modifications to BERT, giving SegaBERT. They find that SegaBERT gives lower validation losses during pre-training, lower language modelling perplexities, and improves upon the GLUE score of BERT. \citet{cheng-etal-2023-mcgill} include POS tags as additional input embeddings during BERT pretraining, which they find to reduce performance on (Super)GLUE \cite{wang2019superglue} and MSGS \cite{warstadt-etal-2020-learning}.

There has also been work which has modified the pretraining objective of transformer models. \citet{yamaguchi-etal-2021-frustratingly} introduce various alternatives to MLM, and pre-train models using them, finding that default MLM is superior in the higher-parameter setting. There have been various works using linguistically-motivated pretraining objectives \cite{zhou2019limit,levine-etal-2020-sensebert}, with the closest to our work being that by \citet{cui2022lert}, who find improved performance through simply adding additional MLM heads for linguistic tasks and summing their losses.

\section{Conclusion}
In this work we investigate whether word boundary information is useful for transformer encoders. In particular, we start with WordPiece$'$, a version of WordPiece modified to remove word boundary information, and show that it leads to more linguistically meaningful tokenisations, as well as improved performance on tasks involving morphologically complex words, whilst having no significant effect on performance for general domain tasks across English and Finnish. We also investigate modifications to the default model architecture which involve incorporating word boundary information, either explicitly (through the input), or implicitly (through the pretraining task), and through pretraining or finetuning alone. Across all models and training scales, we find that these modifications give no substantial improvements in performance, which suggests transformer encoders can perform well without word boundary information, either in the form of prefixes (``\#\#'' or ``\_''), word boundary tokens, word boundary embeddings, or through a modification to the pretraining task.

\section*{Acknowledgements}
This work was partially supported by the CDT in Speech and Language Technologies and their Applications funded by UKRI (grant number EP/S023062/1) and the UK EPSRC grant EP/T02450X/1. We appreciate the helpful comments provided by the reviewers.

\section*{Limitations}
In this work we have only looked at transformer encoder architectures. For encoder-decoder or decoder models, word boundary information needs to be generated in the output -- i.e. WordPiece$'$ is lossy which is problematic for generation. Not including such architectures is a significant limitation of the scope of our work and an important future direction. Despite running pretraining across four scales, we don't look at altering the vocabulary size of our tokenisers, which is another limitation. Whilst we have investigated many approaches to including word boundary information through modified architectures, it is possible that there are alternative approaches which would perform better than these. In addition, whilst we have tried to run experiments on a extensive range of downstream tasks with two languages, it is possible that there are other tasks and languages where the omission of word boundary information would have a significant negative impact on performance.

\bibliography{anthology,custom}

\clearpage
\appendix

\begin{table*}[t!]
\centering
\resizebox{\linewidth}{!}{%
\begin{tabular}{cccccccccccccccc}
 &  \multicolumn{3}{c}{LADEC} &  \multicolumn{3}{c}{MorphoLex}  &  \multicolumn{3}{c}{MorphyNet} &  \multicolumn{3}{c}{DagoBERT} & \multicolumn{3}{c}{MEAN} \\ 
 \cmidrule(lr){2-4}\cmidrule(lr){5-7} \cmidrule(lr){8-10} \cmidrule(lr){11-13} \cmidrule(lr){14-16} 
 & Len & Precis.  & F1 & Len & Precis. & F1 & Len & Precis. & F1 & Len & Precis. & F1 & Len & Precis. & F1 \\
\midrule
WordPiece & 3.34 & 38.0 & 53.3 & 2.91 & 26.0 & 31.4 & 3.43 & 13.2 & 19.7 & 3.47 & 21.9 & 30.7 & 3.29 & 24.8 & 33.8  \\
WordPiece$'$  & 2.66 & \textbf{53.7} & \textbf{67.1} & 2.55 & \textbf{50.0} & \textbf{55.1} & 2.95 & \textbf{25.5} & \textbf{36.1} & 2.85 & \textbf{41.1} & \textbf{52.5} & 2.75 & \textbf{ 42.6} & \textbf{52.7} \\
\bottomrule
\end{tabular}
}
\caption{\label{table:morphology-comparison-english-full} Performance of WordPiece and WordPiece$'$ across four English morphological datasets, showing the average sequence length, precision and F1 score generated following the standard introduced by \citet{creutz2004morpheme}.}
\end{table*}

\begin{table*}[ht!]
\centering  
\resizebox{\linewidth}{!}{%
\renewcommand{\arraystretch}{1.2}
\begin{tabular}{llcccccccccccc}
\toprule
 & Base Dataset & \# Articles (M) & \multicolumn{3}{c}{Examples (M)} & Params (M) & Batch Size & \# GPUs & Steps (k) & \multicolumn{3}{c}{Epochs} & Train Time (h) \\
& & & WP & WP' & WP' spaces & & & & & WP & WP' & WP' Spaces  & \\
\midrule
V Low  & Wikipedia & 0.1 & 1.2 & 1.1 & 1.8 & 5.8 & 1024  & 1 & 25 & 21.5 & 23.0 & 13.9 & 11.0 \\
Low & Wikipedia & 0.5 & 4.1 & 3.8 & 6.3 & 21.2 & 512 & 1  & 50 & 6.3 & 6.7 & 4.1 & 19.0 \\
High  & Wikipedia & 6.5 & 19.8 & 18.5 & 30.2 & 98.2 & 256 & 1 & 400 & 5.2 & 5.5 & 3.4 & 29.2 \\
V High  & C4 & 40 & 88.0 & 82.2 & - & 370.4 & 128 & 4 & 400 & 2.3 & 2.5 & - & 70.9 \\
\bottomrule
\end{tabular}
}
\caption{\label{table:resource_setups_full_english} The four training scales we use to evaluate our models in English.}
\end{table*}

\begin{table*}[ht!]
\centering  
\resizebox{\linewidth}{!}{%
\renewcommand{\arraystretch}{1.2}
\begin{tabular}{llcccccccccc}
\toprule
 & Base Dataset & \# Articles (M) & \multicolumn{2}{c}{Examples (M)} & Params (M) & Batch Size & \# GPUs & Steps (k) & \multicolumn{2}{c}{Epochs} & Train Time (h) \\
& & & WP & WP' & & & & & WP & WP' \\
\midrule
V Low  & Wikipedia & 0.1 & 0.3 & 0.3 & 5.8 & 1024 & 1 & 25 & 78.9 & 84.8 & 4.1 \\
Low & Wikipedia & 2 & 1.0 & 1.0 & 21.2 & 512 & 1 & 50 & 24.7 & 26.6 & 7.8 \\
High  & C4 & 10 & 37.6 & 34.6 & 98.2 & 256 & 1 & 400 & 1.4 & 1.5 & 78.4 \\
\bottomrule
\end{tabular}
}
\caption{\label{table:resource_setups_full_finnish} The three training scales we use to evaluate our models in Finnish.}
\end{table*}

\begin{table}[ht!]
\centering  
\renewcommand{\arraystretch}{1.2}
\begin{tabular}{lccc}
 & Layers & Att. Heads & Embed. Dim. \\
\midrule
V Low  & 2 & 4 & 256\\
Low & 4 & 8 & 512 \\
High  & 12 & 12 & 768 \\
V High  & 26 & 16 & 1024 \\
\bottomrule
\end{tabular}
\caption{\label{table:parameters} Layers, attention heads, and embedding dimension for the four training scales.}
\end{table}

\begin{table*}[ht!]
\centering  
\resizebox{\textwidth}{!}{%
\scriptsize
\renewcommand{\arraystretch}{1.2}
\begin{tabular}{lllll}
\toprule
  & |Train| (k) & |Dev| (k) & Metric & Domain \\
\midrule
CoLA \cite{warstadt2018neural} & 8.5 & 1 & Matthew's Correlation & Books and Journal Articles\\
SST-2 \cite{socher2013recursive}  & 67 & 1 & Accuracy & Film Reviews \\
MRPC \cite{dolan2005automatically} & 3.7 & 0.4 & F1 / Accuracy & Online News \\
STS-B \cite{cer-etal-2017-semeval} & 5.8 & 1.5 & Pearson / Spearman Correlation & Various \\
QQP (\url{ https://quoradata.quora.com/First-Quora-Dataset-Release-Question-Pairs}) & 364 & 40 & F1 / Accuracy & Quora questions \\
MNLI \cite{williams2018broad} & 393 & 9.8 & Accuracy & Various \\
QNLI \cite{rajpurkar2016squad} & 105 & 5.5 & Accuracy & Wikipedia \\
RTE \cite{bentivogli2009fifth} & 2.5 & 0.3 & Accuracy & Wikipedia and News \\
Superbizarre-Arxiv \cite{hofmann-etal-2021-superbizarre} & 58 & 19 & F1 & Arxiv Papers \\
Superbizarre-Reddit \cite{hofmann-etal-2021-superbizarre} & 51 & 17 & F1 & Reddit \\
FLOTA \cite{hofmann-etal-2022-embarrassingly} & 1.2 & 0.4 & F1 & Arxiv Paper Titles \\
FiNER  \cite{ruokolainen2020finnish} & 13.5 & 1.0 & F1 & Online News \\
Eduskunta (\url{https://github.com/aajanki/eduskunta-vkk}) & 49.1 & 3.0 & Accuracy & Parliamentary Questions \\
FinnSentiment \cite{linden2023finnsentiment} & 24.3 & 2.7 & Accuracy & Social Media \\
\bottomrule
\end{tabular}
}
\caption{\label{table:dataset_info} Information for the datasets we use for evaluation. }
\end{table*}

\begin{table*}[ht!]
\resizebox{\textwidth}{!}{%
\begin{tabular}{llrrrrrrrrrrrrrrrrr}
\toprule
 &  & & &  &&  & & & & \multicolumn{2}{l}{mnli} & & & \multicolumn{2}{l}{SB} & \multicolumn{3}{l}{FLOTA} \\
 &  & \multicolumn{1}{l}{conll} & \multicolumn{1}{l}{ncbi} & \multicolumn{1}{l}{wnut17} & \multicolumn{1}{l}{cola} & \multicolumn{1}{l}{sst2} & \multicolumn{1}{l}{mrpc} & \multicolumn{1}{l}{stsb} & \multicolumn{1}{l}{qqp} & \multicolumn{1}{l}{m} & \multicolumn{1}{l}{mm} & \multicolumn{1}{l}{qnli} & \multicolumn{1}{l}{rte} & \multicolumn{1}{l}{A} & \multicolumn{1}{l}{R} & \multicolumn{1}{l}{CS} & \multicolumn{1}{l}{M} & \multicolumn{1}{l}{P} \\
 \midrule
\textbf{V Low} & WP & 79.5 & 59.5 & 24.0 & 6.9 & 79.7 & 76.0 & 15.7 & 74.0 & 59.9 & 61.1 & 64.6 & 54.2 & 66.1 & 63.9 & 20.8 & 16.9 & 20.8 \\
 & WP$'$ & 80.2 & 60.4 & 20.4 & 3.5 & 80.8 & 76.2 & 15.8 & 77.4 & 63.8 & 65.1 & 67.3 & 56.0 & 68.3 & 65.4 & 23.0 & 23.1 & 24.7 \\
 & WP$'$ extra loss & 80.1 & 61.3 & 23.0 & 6.7 & 80.6 & 75.6 & 16.0 & 76.7 & 63.3 & 64.9 & 65.7 & 55.3 & 68.4 & 65.4 & 21.6 & 20.3 & 28.7 \\
 & WP$'$ binary & 79.7 & 60.8 & 22.6 & 7.4 & 82.7 & 76.0 & 15.2 & 74.7 & 62.5 & 63.6 & 63.6 & 55.4 & 68.3 & 65.4 & 24.2 & 21.8 & 27.4 \\
 & WP$'$ word pos & 80.3 & 60.5 & 23.9 & 7.7 & 81.4 & 76.1 & 20.6 & 78.6 & 63.7 & 65.1 & 68.0 & 53.8 & 68.3 & 65.4 & 21.7 & 19.6 & 25.6 \\
 & WP$'$ subword pos & 80.9 & 62.5 & 21.6 & 5.7 & 82.8 & 75.9 & 13.1 & 74.2 & 63.0 & 64.3 & 65.0 & 56.0 & 68.6 & 65.3 & 26.0 & 22.3 & 24.6 \\
 & WP$'$ spaces & 78.0 & 58.8 & 20.4 & 7.3 & 80.6 & 76.5 & 14.4 & 75.6 & 62.0 & 62.6 & 64.9 & 53.6 & 67.8 & 65.3 & 20.0 & 19.9 & 23.2 \\
 & WP$'$ f/t binary & 80.3 & 60.3 & 20.1 & 4.7 & 81.0 & 75.5 & 16.0 & 77.3 & 63.8 & 65.0 & 66.4 & 55.7 & 68.4 & 65.4 & 22.1 & 22.1 & 25.6 \\
 & WP$'$ f/t spaces & 80.2 & 60.4 & 20.4 & 1.9 & 81.4 & 76.5 & 14.0 & 74.4 & 63.3 & 64.6 & 65.2 & 54.9 & - & - & 22.3 & 21.9 & 26.0 \\
\midrule
\textbf{Low} & WP & 89.3 & 78.0 & 39.5 & 11.9 & 85.0 & 78.2 & 66.1 & 85.2 & 71.9 & 72.2 & 82.0 & 56.7 & 68.1 & 64.4 & 30.3 & 28.7 & 34.7 \\
 & WP$'$ & 89.9 & 77.0 & 37.3 & 16.9 & 85.3 & 79.0 & 78.0 & 85.6 & 72.2 & 72.7 & 81.6 & 56.6 & 69.3 & 66.0 & 44.1 & 38.7 & 46.6 \\
 & WP$'$ extra loss & 90.7 & 77.6 & 39.3 & 15.5 & 84.6 & 77.3 & 75.2 & 85.2 & 72.6 & 73.1 & 81.2 & 56.3 & 69.2 & 65.9 & 46.3 & 41.0 & 48.1 \\
 & WP$'$ binary & 89.9 & 77.5 & 37.2 & 18.5 & 87.1 & 77.0 & 76.3 & 85.3 & 73.4 & 73.4 & 82.7 & 57.3 & 69.1 & 66.0 & 44.6 & 41.7 & 47.3 \\
 & WP$'$ word pos & 89.7 & 77.1 & 38.3 & 16.3 & 84.2 & 78.0 & 76.4 & 84.8 & 71.4 & 72.0 & 81.8 & 57.5 & 69.2 & 66.0 & 42.9 & 39.4 & 47.4 \\
 & WP$'$ subword pos & 90.0 & 77.0 & 37.1 & 18.4 & 86.3 & 78.9 & 77.5 & 85.6 & 72.8 & 73.0 & 82.5 & 58.1 & 69.5 & 66.0 & 40.2 & 33.6 & 40.8 \\
 & WP$'$ spaces & 89.1 & 76.3 & 37.5 & 16.3 & 84.3 & 76.0 & 74.3 & 85.0 & 72.0 & 72.2 & 80.0 & 58.1 & 69.2 & 65.7 & 43.2 & 40.4 & 47.0 \\
 & WP$'$ f/t binary & 90.0 & 77.3 & 38.6 & 16.0 & 84.9 & 79.1 & 77.5 & 85.5 & 72.4 & 73.0 & 82.3 & 58.6 & 69.3 & 65.8 & 44.0 & 40.6 & 46.1 \\
 & WP$'$ f/t spaces & 90.0 & 76.9 & 38.1 & 16.3 & 84.3 & 78.2 & 79.6 & 85.3 & 71.9 & 72.9 & 81.4 & 57.1 & - & - & 45.8 & 39.3 & 46.0 \\
\midrule
\textbf{High} & WP & 95.0 & 83.7 & 52.1 & 34.7 & 90.0 & 87.2 & 85.6 & 88.9 & 80.3 & 80.4 & 89.1 & 65.1 & 69.5 & 65.2 & 51.6 & 47.6 & 52.0 \\
 & WP$'$ & 94.9 & 83.7 & 48.6 & 40.2 & 90.8 & 87.3 & 85.7 & 88.6 & 79.9 & 80.0 & 87.1 & 62.8 & 70.3 & 66.5 & 53.2 & 49.8 & 53.8 \\
 & WP$'$ extra loss & 94.9 & 83.2 & 48.7 & 34.6 & 90.4 & 87.2 & 85.7 & 88.5 & 79.5 & 80.0 & 88.5 & 65.6 & 70.6 & 66.1 & 53.5 & 48.5 & 53.3 \\
 & WP$'$ binary & 94.6 & 84.0 & 47.4 & 40.4 & 90.5 & 87.6 & 85.7 & 88.7 & 79.9 & 80.3 & 88.5 & 64.3 & 70.1 & 66.2 & 52.0 & 48.1 & 53.0 \\
 & WP$'$ word pos & 94.6 & 83.1 & 48.5 & 40.0 & 90.7 & 88.2 & 86.3 & 88.7 & 80.5 & 80.4 & 88.2 & 66.1 & 70.3 & 66.5 & 51.9 & 49.0 & 54.7 \\
 & WP$'$ subword pos & 94.4 & 83.7 & 48.1 & 38.4 & 90.4 & 86.5 & 85.7 & 88.8 & 80.4 & 80.6 & 87.9 & 63.8 & 70.0 & 66.4 & 47.3 & 46.3 & 51.0 \\
 & WP$'$ spaces & 93.8 & 83.8 & 44.6 & 38.0 & 90.2 & 86.6 & 84.9 & 88.3 & 79.3 & 79.5 & 86.4 & 64.3 & 70.3 & 66.2 & 54.0 & 49.6 & 53.2 \\
 & WP$'$ f/t binary & 94.8 & 83.3 & 48.2 & 39.0 & 90.9 & 87.1 & 85.8 & 88.5 & 79.9 & 80.0 & 87.0 & 61.9 & 70.1 & 66.0 & 51.6 & 51.6 & 54.7 \\
 & WP$'$ f/t WB tokens & 94.9 & 83.7 & 48.6 & 36.3 & 90.5 & 87.1 & 85.7 & 88.4 & 80.1 & 80.6 & 86.7 & 63.3 & - & - & 52.9 & 49.9 & 54.7 \\
 \midrule
 \textbf{V High} & WP & 95.6 & 86.1 & 62.9 & 61.3 & 92.3 & 89.2 & 89.0 & 89.9 & 85.6 & 85.7 & 91.2 & 63.9 & 70.6 & 66.5 & 59.2 & 51.4 & 54.3 \\
 & WP$'$ & 95.7 & 86.5 & 62.2 & 61.3 & 93.1 & 90.9 & 89.4 & 90.0 & 85.2 & 85.3 & 90.9 & 67.1 & 71.6 & 67.4 & 60.4 & 49.4 & 55.6 \\
 \bottomrule
\end{tabular}
}
\caption{\label{table:full_results} Full English results across all datasets, training scales, and models. }
\end{table*}

\begin{table}[ht!]
\resizebox{\columnwidth}{!}{%
\begin{tabular}{llrrr}
\toprule
 &  & FiNER & Eduskunta & FinnSentiment \\
 \midrule
\textbf{V Low} & WP & 72.2 &64.6 & 81.6 \\
 & WP$'$ & 73.0 & 65.2 & 80.9 \\
\midrule
\textbf{Low} & WP & 84.2 & 71.3 & 86.3 \\
 & WP$'$ & 85.0 & 71.1 & 86.8 \\
\midrule
\textbf{High} & WP & 89.9 & 75.9 & 91.3 \\
 & WP$'$ & 89.8 & 75.3 & 92.9 \\
 \bottomrule
\end{tabular}
}
\caption{\label{table:full_results_finnish} Full Finnish results across all datasets, training scales, and models. }
\end{table}

\begin{figure*}[ht!]

    \begin{subfigure}[t]{0.32\textwidth}
    \begin{center}
        \resizebox{.95\linewidth}{!}{\includegraphics{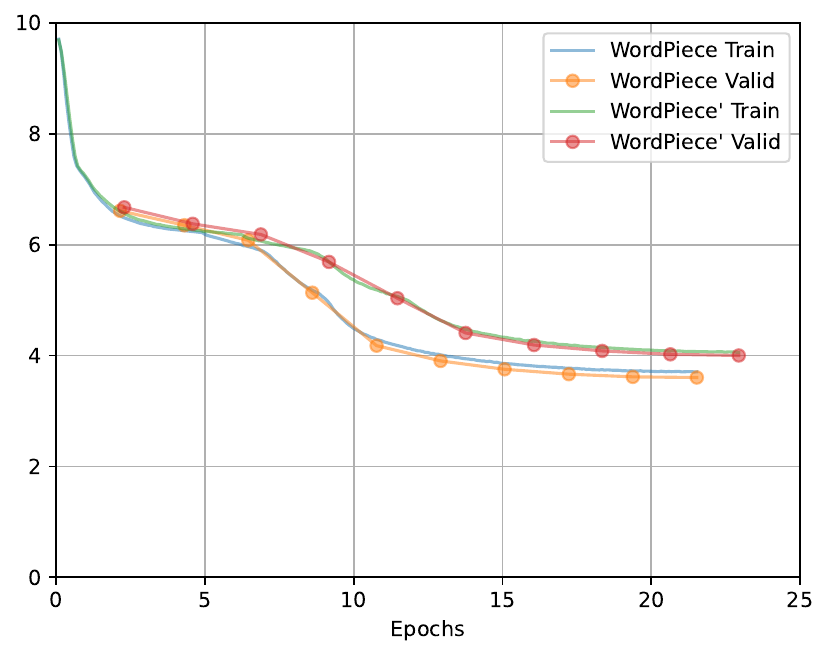}}
    \end{center}
    \caption{V Low}
    \end{subfigure}
    \begin{subfigure}[t]{0.32\textwidth}
    \begin{center}
       \resizebox{.95\linewidth}{!}{\includegraphics{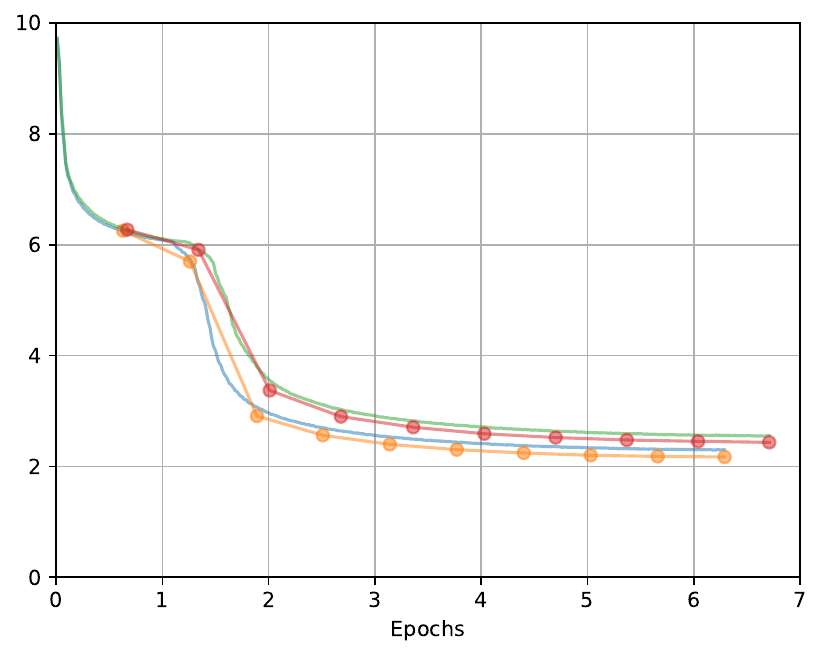}}
    \end{center}
    \caption{Low}
    \end{subfigure}
    \begin{subfigure}[t]{0.32\textwidth}
    \begin{center}
        \resizebox{.95\linewidth}{!}{\includegraphics{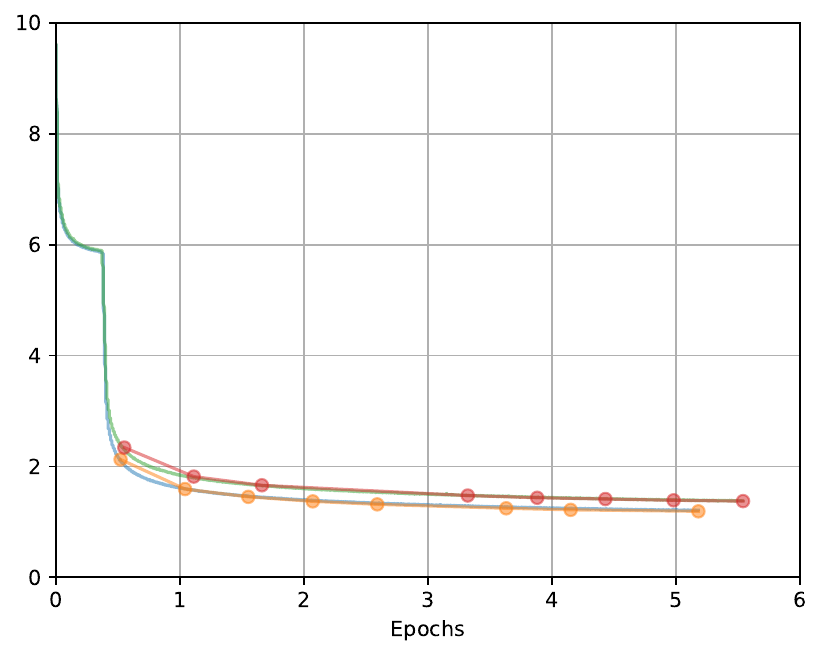}}
    \end{center}
    \caption{High}
    \end{subfigure}
    
    \caption{\label{figure:loss_curves_english} Training and valid losses for WordPiece and WordPiece$'$ across three training scales for English.}
\end{figure*}

\begin{figure*}[ht!]

    \begin{subfigure}[t]{0.32\textwidth}
    \begin{center}
        \resizebox{.95\linewidth}{!}{\includegraphics{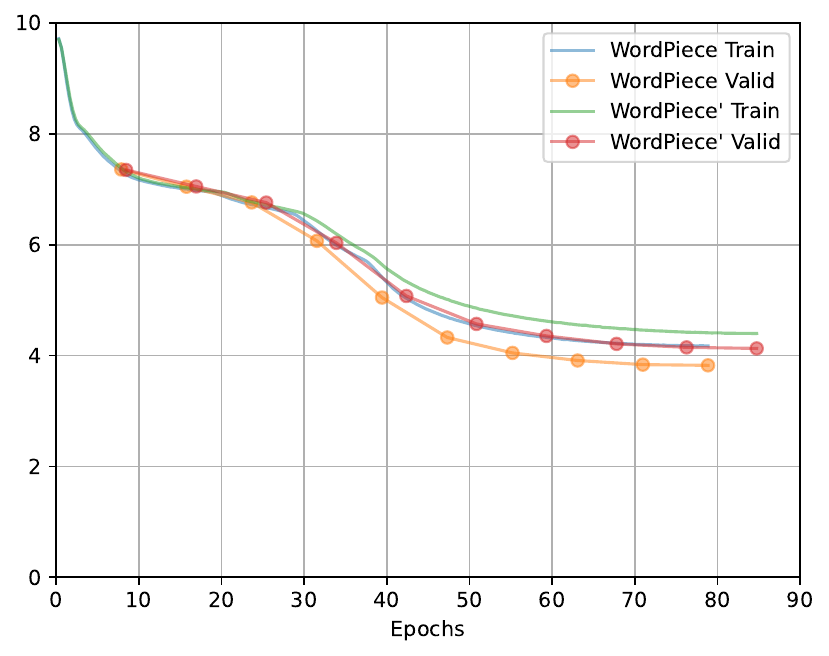}}
    \end{center}
    \caption{V Low}
    \end{subfigure}
    \begin{subfigure}[t]{0.32\textwidth}
    \begin{center}
       \resizebox{.95\linewidth}{!}{\includegraphics{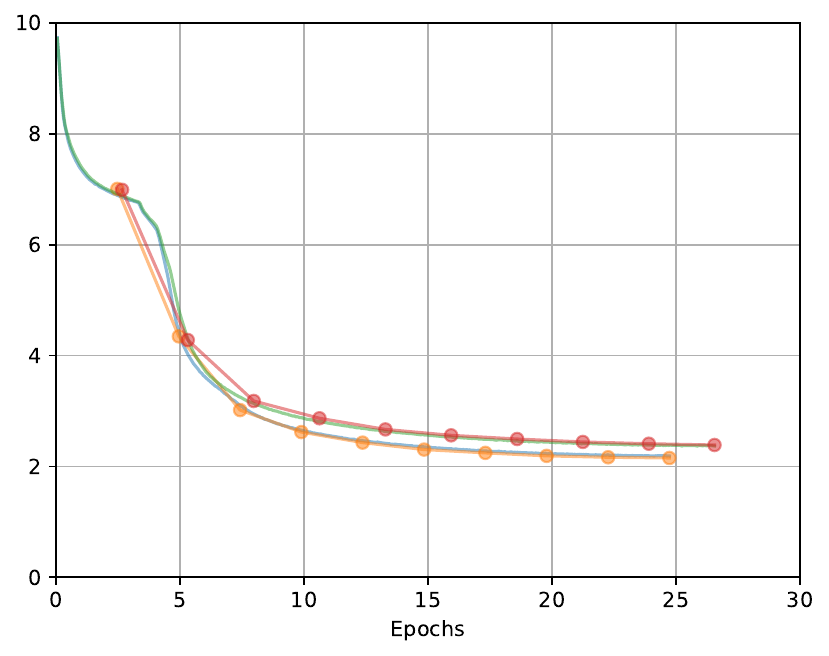}}
    \end{center}
    \caption{Low}
    \end{subfigure}
    \begin{subfigure}[t]{0.32\textwidth}
    \begin{center}
        \resizebox{.95\linewidth}{!}{\includegraphics{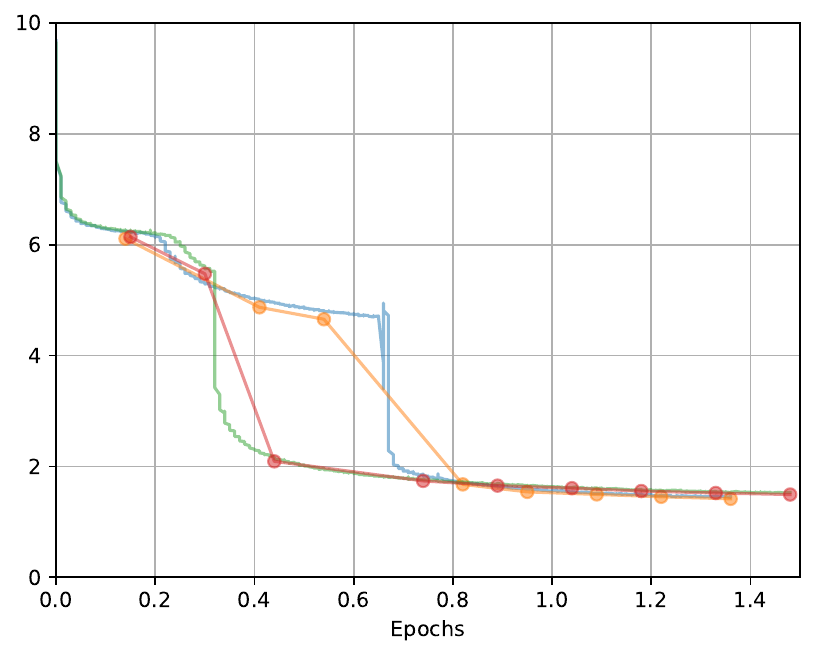}}
    \end{center}
    \caption{High}
    \end{subfigure}
    
    \caption{\label{figure:loss_curves_finnish} Training and valid losses for WordPiece and WordPiece$'$ across three training scales for Finnish.}
\end{figure*}

\begin{figure*}[ht!]

    \begin{subfigure}[t]{0.24\textwidth}
    \begin{center}
        \resizebox{.95\linewidth}{!}{\includegraphics{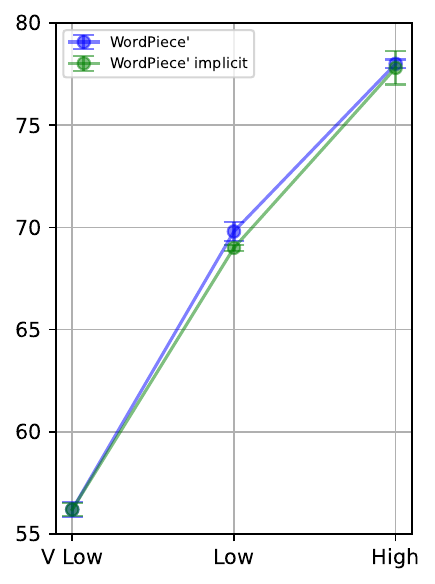}}
    \end{center}
    \caption{GLUE}
    \end{subfigure}
    \begin{subfigure}[t]{0.24\textwidth}
    \begin{center}
       \resizebox{.95\linewidth}{!}{\includegraphics{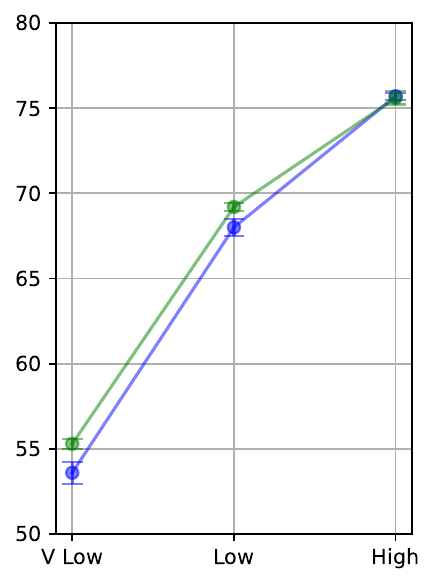}}
    \end{center}
    \caption{NER}
    \end{subfigure}
    \begin{subfigure}[t]{0.24\textwidth}
    \begin{center}
        \resizebox{.95\linewidth}{!}{\includegraphics{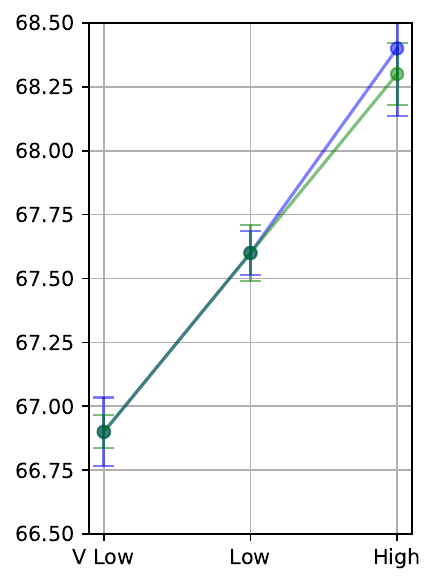}}
    \end{center}
    \caption{Superbizarre}
    \end{subfigure}
    \begin{subfigure}[t]{0.24\textwidth}
    \begin{center}
        \resizebox{.95\linewidth}{!}{\includegraphics{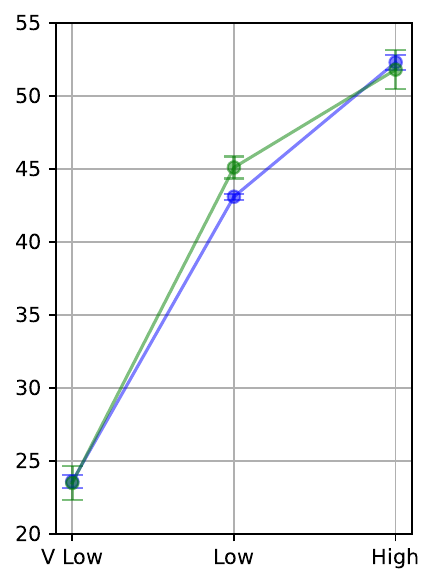}}
    \end{center}
    \caption{FLOTA}
    \end{subfigure}
    
    \caption{\label{figure:results_wordpiece_extra_loss} Results for WordPiece$'$ and WordPiece$'$ implicit.}
\end{figure*}

\begin{figure*}[t!]

    \begin{subfigure}[t]{0.24\textwidth}
    \begin{center}
        \resizebox{.99\linewidth}{!}{\includegraphics{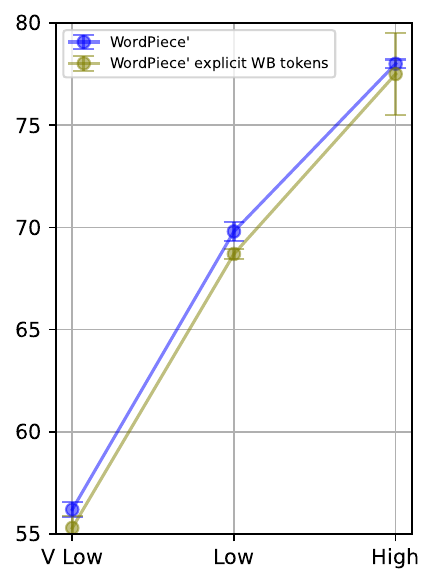}}
    \end{center}
    \caption{GLUE}
    \end{subfigure}
    \begin{subfigure}[t]{0.24\textwidth}
    \begin{center}
       \resizebox{.99\linewidth}{!}{\includegraphics{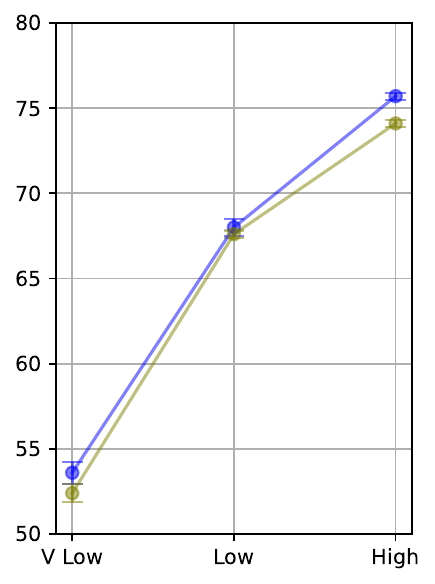}}
    \end{center}
    \caption{NER}
    \end{subfigure}
    \begin{subfigure}[t]{0.24\textwidth}
    \begin{center}
        \resizebox{.99\linewidth}{!}{\includegraphics{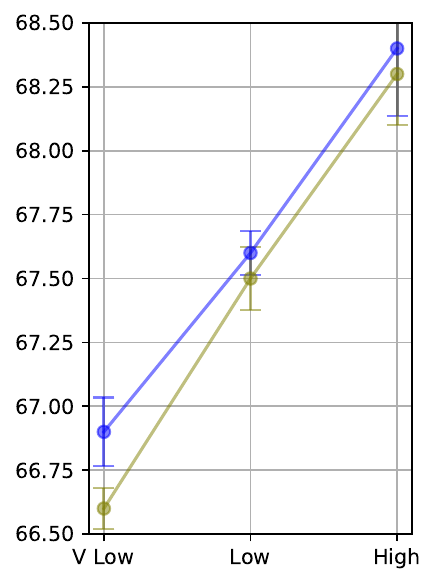}}
    \end{center}
    \caption{Superbizarre}
    \end{subfigure}
    \begin{subfigure}[t]{0.24\textwidth}
    \begin{center}
        \resizebox{.99\linewidth}{!}{\includegraphics{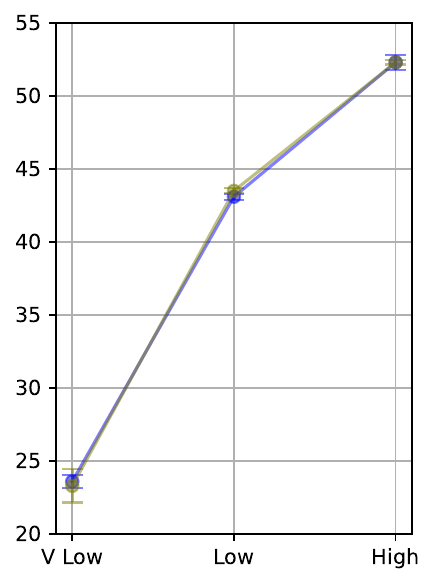}}
    \end{center}
    \caption{FLOTA}
    \end{subfigure}
    
    \caption{\label{figure:results_wordpiece_prime_spaces} Results for WordPiece$'$ and WordPiece$'$ explicit with word boundary tokens.}
\end{figure*}

\begin{figure*}

    \begin{subfigure}[t]{0.24\textwidth}
    \begin{center}
        \resizebox{.99\linewidth}{!}{\includegraphics{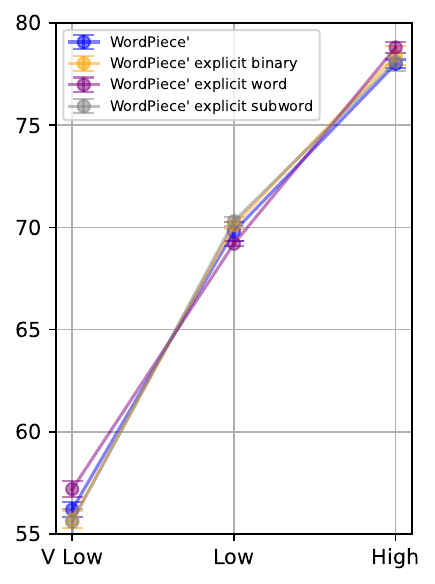}}
    \end{center}
    \caption{GLUE}
    \end{subfigure}
    \begin{subfigure}[t]{0.24\textwidth}
    \begin{center}
       \resizebox{.99\linewidth}{!}{\includegraphics{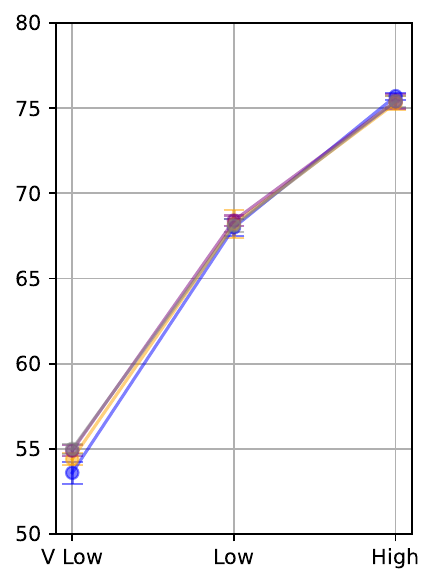}}
    \end{center}
    \caption{NER}
    \end{subfigure}
    \begin{subfigure}[t]{0.24\textwidth}
    \begin{center}
        \resizebox{.99\linewidth}{!}{\includegraphics{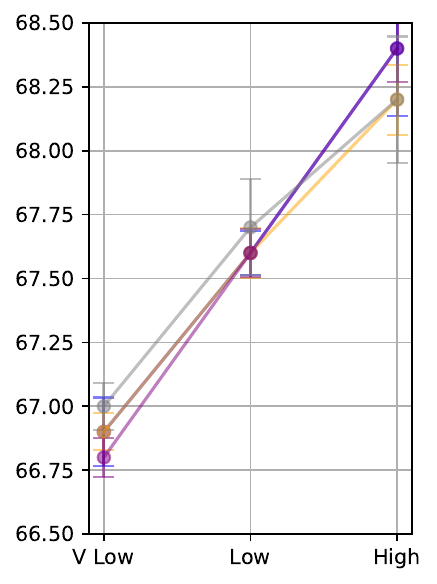}}
    \end{center}
    \caption{Superbizarre}
    \end{subfigure}
    \begin{subfigure}[t]{0.24\textwidth}
    \begin{center}
        \resizebox{.99\linewidth}{!}{\includegraphics{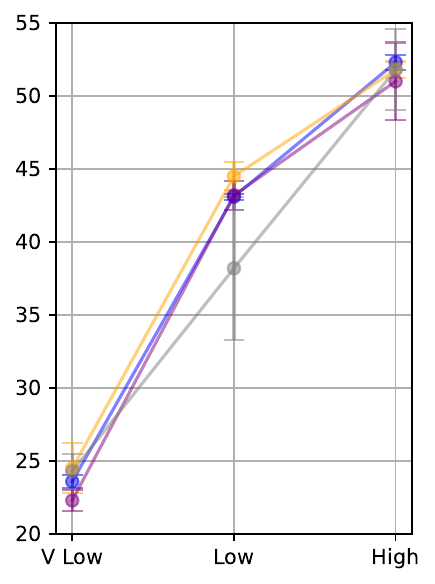}}
    \end{center}
    \caption{FLOTA}
    \end{subfigure}
    
    \caption{\label{figure:results_wordpiece_prime_wb_embeddings} Results for WordPiece$'$ and WordPiece$'$ explicit with word boundary embeddings.}
\end{figure*}

\begin{figure*}

    \begin{subfigure}[t]{0.24\textwidth}
    \begin{center}
        \resizebox{.99\linewidth}{!}{\includegraphics{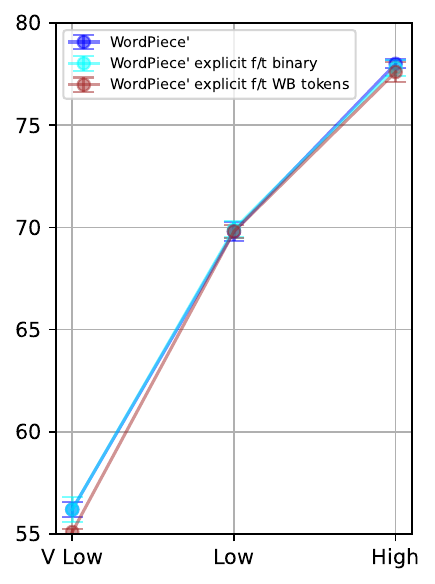}}
    \end{center}
    \caption{GLUE}
    \end{subfigure}
    \begin{subfigure}[t]{0.24\textwidth}
    \begin{center}
       \resizebox{.99\linewidth}{!}{\includegraphics{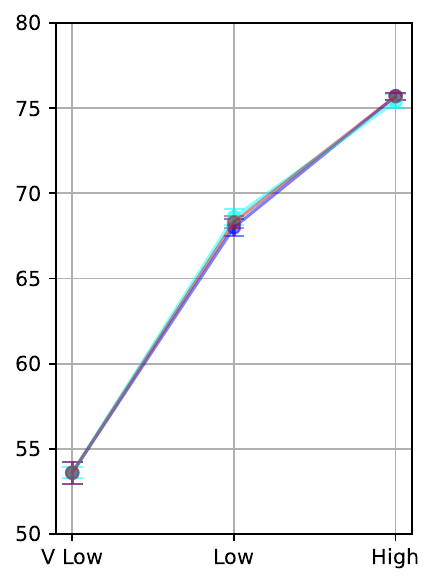}}
    \end{center}
    \caption{NER}
    \end{subfigure}
    \begin{subfigure}[t]{0.24\textwidth}
    \begin{center}
        \resizebox{.99\linewidth}{!}{\includegraphics{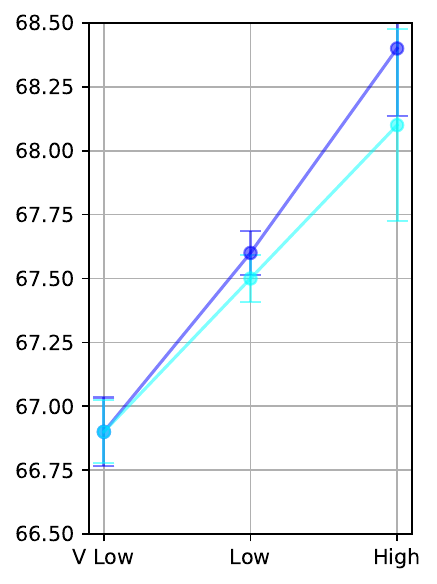}}
    \end{center}
    \caption{Superbizarre}
    \end{subfigure}
    \begin{subfigure}[t]{0.24\textwidth}
    \begin{center}
        \resizebox{.99\linewidth}{!}{\includegraphics{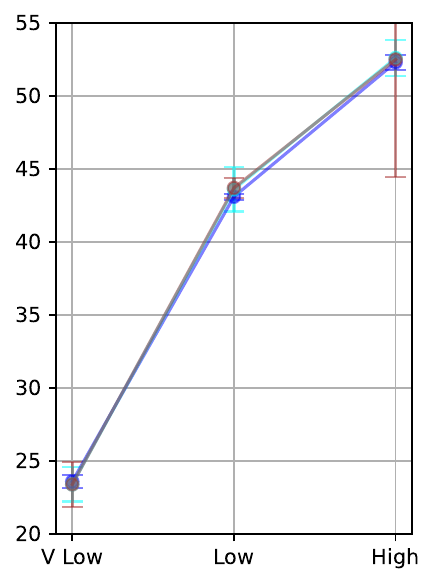}}
    \end{center}
    \caption{FLOTA}
    \end{subfigure}
    
    \caption{\label{figure:results_wordpiece_prime_ft} Results for WordPiece$'$ and WordPiece$'$ finetuned with either word boundary tokens or binary index word boundary embeddings.}
\end{figure*}

\begin{figure}
\begin{center}
    \resizebox{.99\linewidth}{!}{\includegraphics{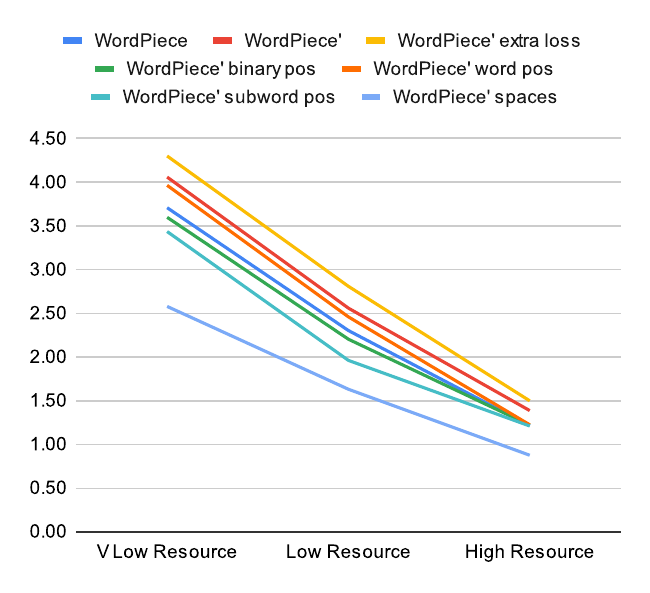}}
\end{center}
\caption{\label{figure:losses} Pretraining MLM losses for all English models across three training scales, averaged across the last 100 steps.}
\end{figure}

\begin{figure}
\def\svgwidth{\columnwidth}
\centering
\resizebox{.99\linewidth}{!}{\includegraphics{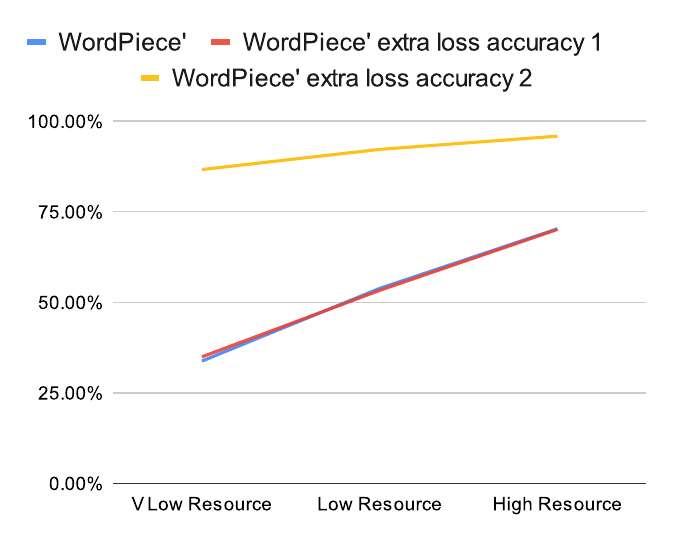}}
\caption{\label{fig:wordpiece_prime_eval_losses} English pretraining evaluation accuracies for WordPiece$'$ and the two MLM heads for WordPiece$'$ extra loss.}
\end{figure}

\begin{figure*}

    \begin{subfigure}[t]{0.24\textwidth}
    \begin{center}
        \resizebox{.99\linewidth}{!}{\includegraphics{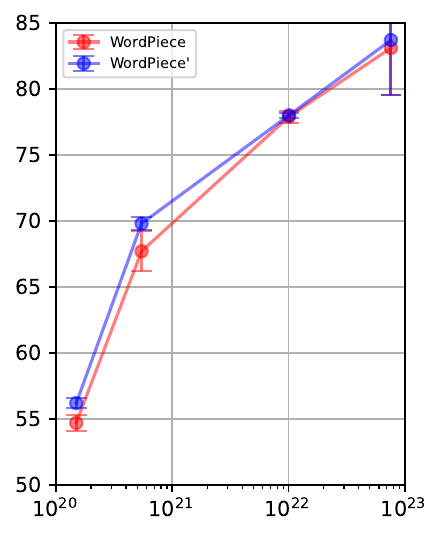}}
    \end{center}
    \caption{GLUE}
    \end{subfigure}
    \begin{subfigure}[t]{0.24\textwidth}
    \begin{center}
       \resizebox{.99\linewidth}{!}{\includegraphics{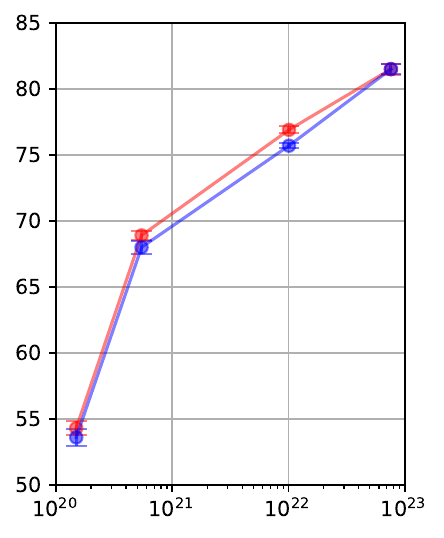}}
    \end{center}
    \caption{NER}
    \end{subfigure}
    \begin{subfigure}[t]{0.24\textwidth}
    \begin{center}
        \resizebox{.99\linewidth}{!}{\includegraphics{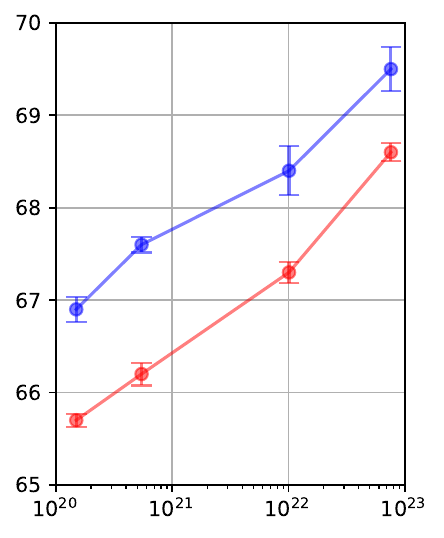}}
    \end{center}
    \caption{\label{figure:log_results_wordpiece_wordpiece_prime_superbizarre}Superbizarre}
    \end{subfigure}
    \begin{subfigure}[t]{0.24\textwidth}
    \begin{center}
        \resizebox{.99\linewidth}{!}{\includegraphics{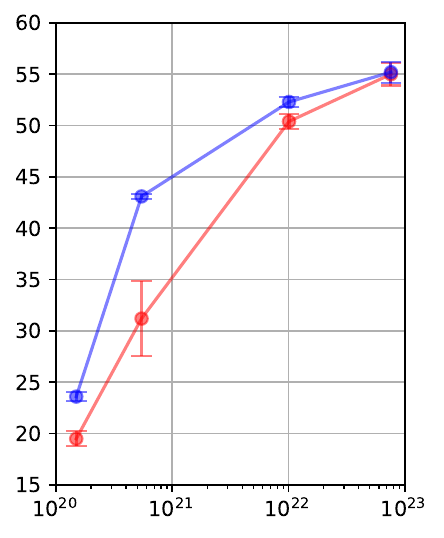}}
    \end{center}
    \caption{FLOTA}
    \end{subfigure}
    
    \caption{\label{figure:log_results_wordpiece_wordpiece_prime} Results for WordPiece and WordPiece$'$ with log training scale on the x-axis.}
\end{figure*}

\begin{figure*}

    \begin{subfigure}[t]{0.24\textwidth}
    \begin{center}
        \resizebox{.95\linewidth}{!}{\includegraphics{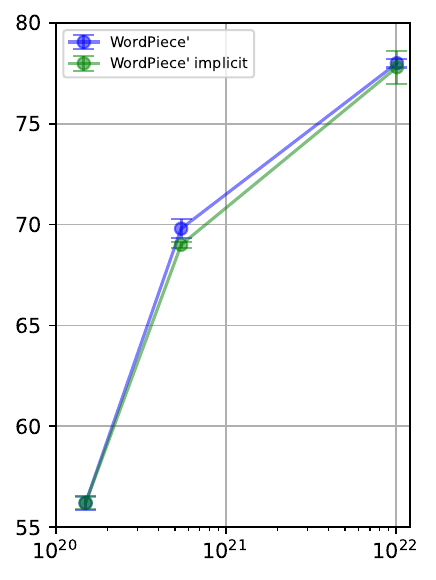}}
    \end{center}
    \caption{GLUE}
    \end{subfigure}
    \begin{subfigure}[t]{0.24\textwidth}
    \begin{center}
       \resizebox{.95\linewidth}{!}{\includegraphics{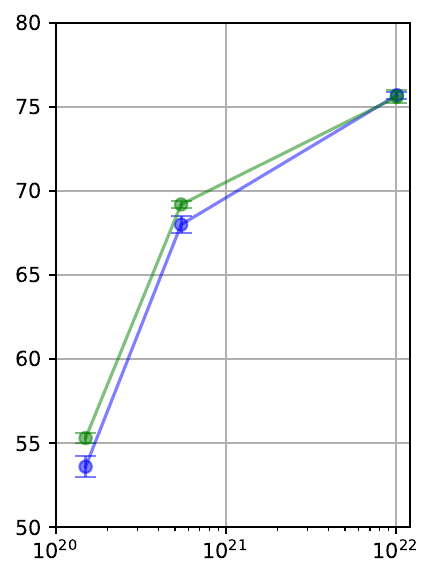}}
    \end{center}
    \caption{NER}
    \end{subfigure}
    \begin{subfigure}[t]{0.24\textwidth}
    \begin{center}
        \resizebox{.95\linewidth}{!}{\includegraphics{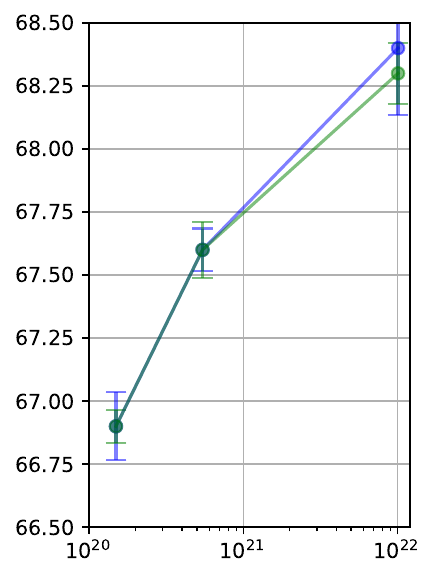}}
    \end{center}
    \caption{Superbizarre}
    \end{subfigure}
    \begin{subfigure}[t]{0.24\textwidth}
    \begin{center}
        \resizebox{.95\linewidth}{!}{\includegraphics{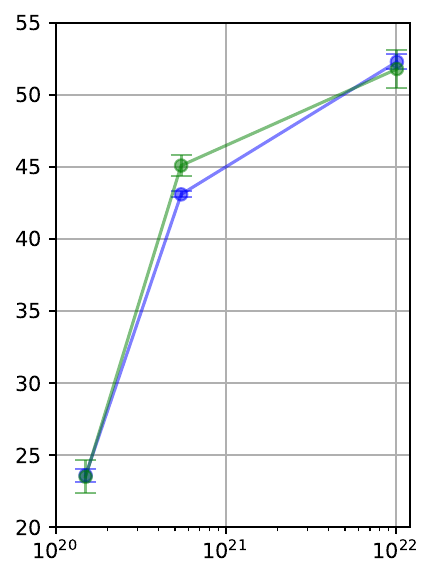}}
    \end{center}
    \caption{FLOTA}
    \end{subfigure}
    
    \caption{\label{figure:log_results_wordpiece_extra_loss} Results for WordPiece$'$ and WordPiece$'$ implicit with log training scale on the x-axis.}
\end{figure*}

\begin{figure*}

    \begin{subfigure}[t]{0.24\textwidth}
    \begin{center}
        \resizebox{.99\linewidth}{!}{\includegraphics{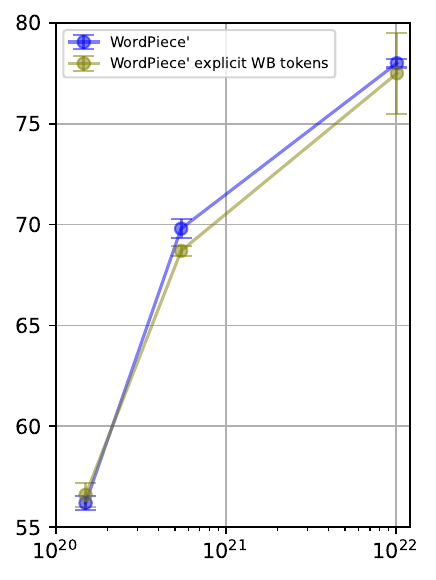}}
    \end{center}
    \caption{GLUE}
    \end{subfigure}
    \begin{subfigure}[t]{0.24\textwidth}
    \begin{center}
       \resizebox{.99\linewidth}{!}{\includegraphics{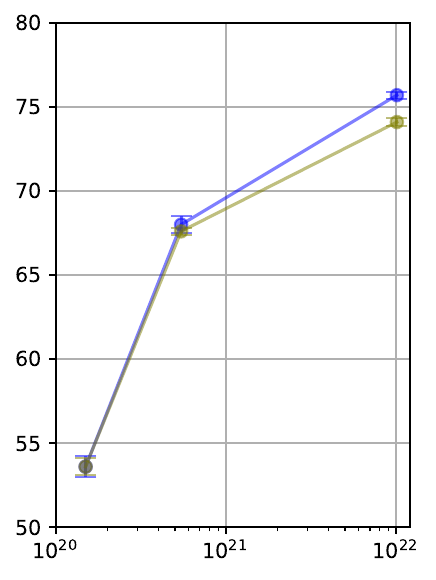}}
    \end{center}
    \caption{NER}
    \end{subfigure}
    \begin{subfigure}[t]{0.24\textwidth}
    \begin{center}
        \resizebox{.99\linewidth}{!}{\includegraphics{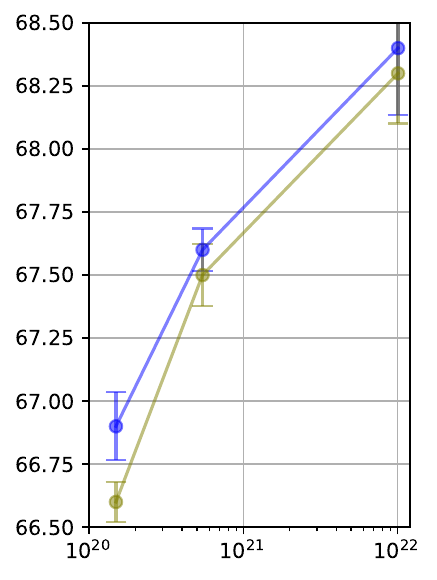}}
    \end{center}
    \caption{Superbizarre}
    \end{subfigure}
    \begin{subfigure}[t]{0.24\textwidth}
    \begin{center}
        \resizebox{.99\linewidth}{!}{\includegraphics{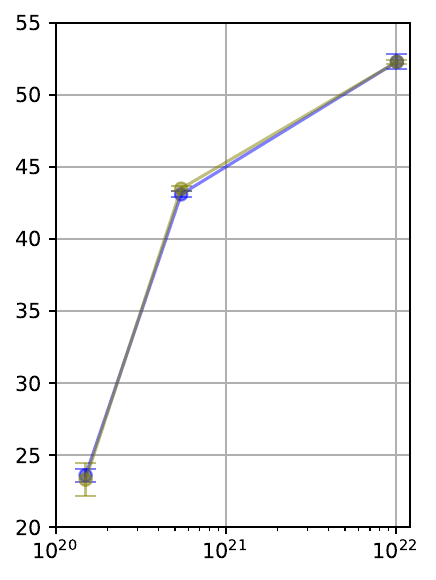}}
    \end{center}
    \caption{FLOTA}
    \end{subfigure}
    
    \caption{\label{figure:log_results_wordpiece_prime_spaces} Results for WordPiece$'$ and WordPiece$'$ explicit with word boundary tokens with log training scale on the x-axis.}
\end{figure*}

\begin{figure*}

    \begin{subfigure}[t]{0.24\textwidth}
    \begin{center}
        \resizebox{.99\linewidth}{!}{\includegraphics{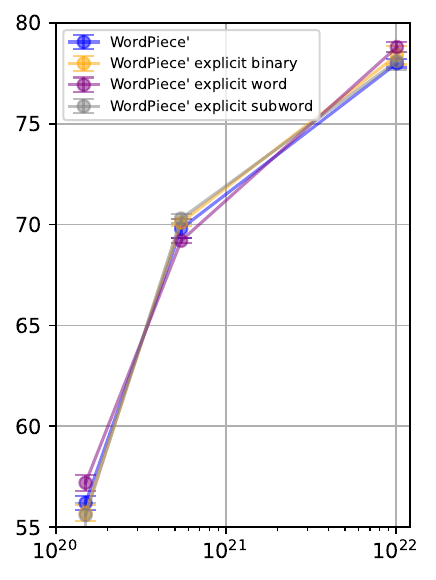}}
    \end{center}
    \caption{GLUE}
    \end{subfigure}
    \begin{subfigure}[t]{0.24\textwidth}
    \begin{center}
       \resizebox{.99\linewidth}{!}{\includegraphics{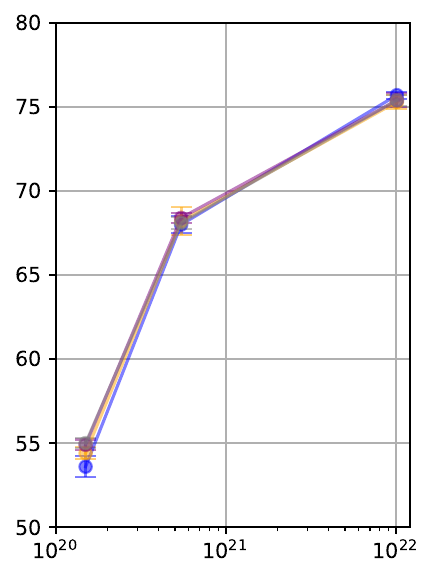}}
    \end{center}
    \caption{NER}
    \end{subfigure}
    \begin{subfigure}[t]{0.24\textwidth}
    \begin{center}
        \resizebox{.99\linewidth}{!}{\includegraphics{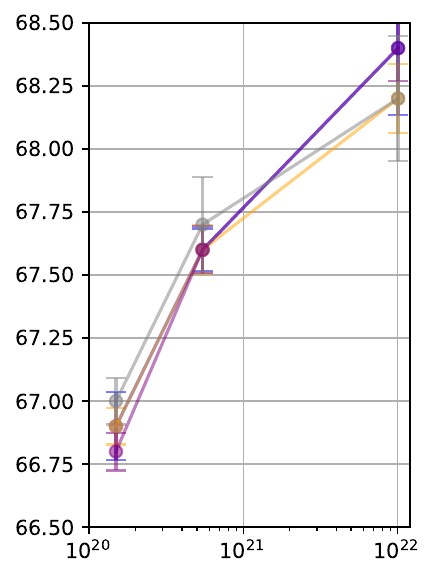}}
    \end{center}
    \caption{Superbizarre}
    \end{subfigure}
    \begin{subfigure}[t]{0.24\textwidth}
    \begin{center}
        \resizebox{.99\linewidth}{!}{\includegraphics{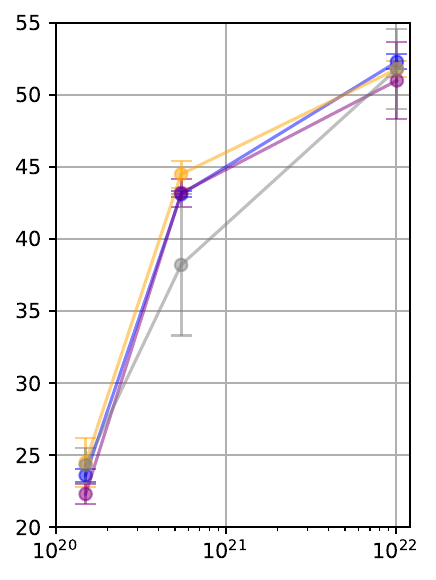}}
    \end{center}
    \caption{FLOTA}
    \end{subfigure}
    
    \caption{\label{figure:log_results_wordpiece_prime_wb_embeddings} Results for WordPiece$'$ and WordPiece$'$ explicit with word boundary embeddings with log training scale on the x-axis.}
\end{figure*}

\begin{figure*}

    \begin{subfigure}[t]{0.24\textwidth}
    \begin{center}
        \resizebox{.99\linewidth}{!}{\includegraphics{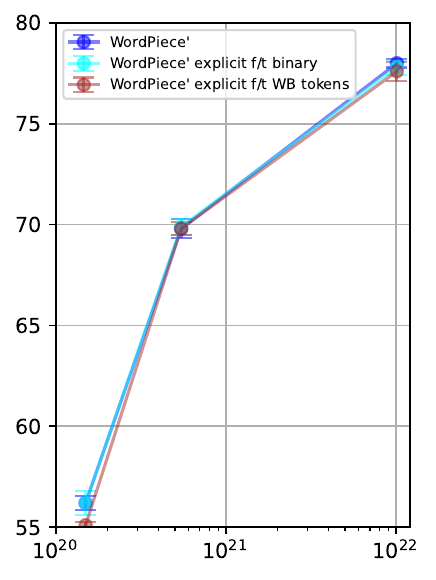}}
    \end{center}
    \caption{GLUE}
    \end{subfigure}
    \begin{subfigure}[t]{0.24\textwidth}
    \begin{center}
       \resizebox{.99\linewidth}{!}{\includegraphics{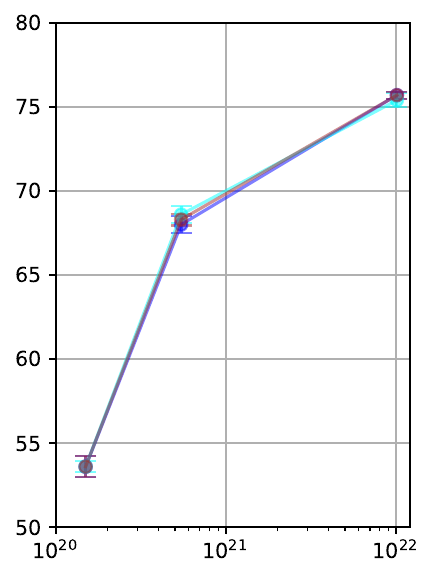}}
    \end{center}
    \caption{NER}
    \end{subfigure}
    \begin{subfigure}[t]{0.24\textwidth}
    \begin{center}
        \resizebox{.99\linewidth}{!}{\includegraphics{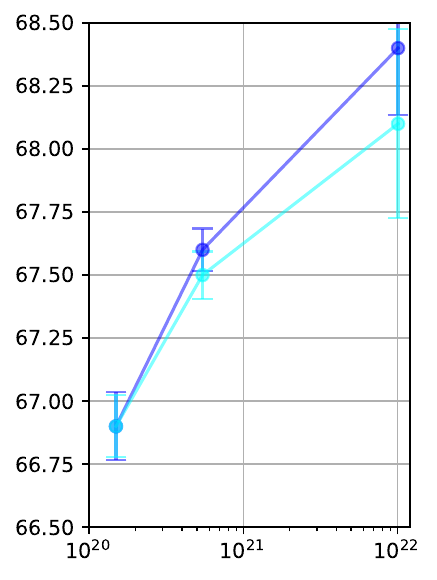}}
    \end{center}
    \caption{Superbizarre}
    \end{subfigure}
    \begin{subfigure}[t]{0.24\textwidth}
    \begin{center}
        \resizebox{.99\linewidth}{!}{\includegraphics{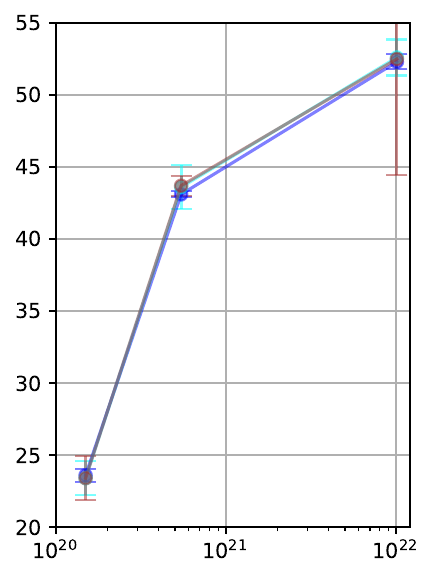}}
    \end{center}
    \caption{FLOTA}
    \end{subfigure}
    
    \caption{\label{figure:log_results_wordpiece_prime_ft} Results for WordPiece$'$ and WordPiece$'$ finetuned with either word boundary tokens or binary index word boundary embeddings with log training scale on the x-axis.}
\end{figure*}

\end{document}